\documentclass{article} % For LaTeX2e
\usepackage{iclr2027_conference,times}

\usepackage{amsmath,amsfonts,bm}

\def\eqref#1{Eq.~(\ref{#1})}
\def\1{\bm{1}}

\DeclareMathAlphabet{\mathsfit}{\encodingdefault}{\sfdefault}{m}{sl}
\SetMathAlphabet{\mathsfit}{bold}{\encodingdefault}{\sfdefault}{bx}{n}

\newcommand{\E}{\mathbb{E}}

\newcommand{\Cov}{\mathrm{Cov}}
\definecolor{celadon}{rgb}{0.75, 0.88, 0.85}
\usepackage{graphicx}
\usepackage{booktabs}
\usepackage{wrapfig}
\usepackage[title]{appendix}
\usepackage{tabularx}
\usepackage{adjustbox}
\usepackage{multicol}
\usepackage{multirow}
\usepackage{caption}

\usepackage{subcaption}
\usepackage{array}

\usepackage{pifont}

\usepackage[table]{xcolor}
\definecolor{lncolor}{HTML}{FEE4C4}
\definecolor{baseline}{HTML}{E0E4E8}
\definecolor{ppink}{rgb}{0.98, 0.575, 0.89}

\newcommand{\hlrow}{\rowcolor{celadon}}
\newcommand{\graya}{\rowcolor{gray!20}}

\usepackage{scalerel}
\newcommand{\shortpara}{\stretchrel*{\parallel}{\perp}}

\DeclareMathOperator{\mupar}{\mu_i^{\shortpara}}
\DeclareMathOperator{\muort}{\mu_i^{\perp}}

\newcommand{\minisection}[1]{\vspace{3pt} \noindent {\bf #1}}

\usepackage{url}

\usepackage[pagebackref]{hyperref}

\usepackage[capitalize,noabbrev]{cleveref}

\title{Project and Mix: Task-Semantic Prototypes for Few-Shot Image Classification}

\author{Dipam Goswami\textsuperscript{1}\thanks{Equal contributions} \quad Simone Magistri\textsuperscript{2}\footnotemark[1] \quad Gido M.\ van de Ven\textsuperscript{3} \quad Bartłomiej Twardowski\textsuperscript{1,4,6} \\
\vspace{2mm}
\textbf{Andrew D. Bagdanov\textsuperscript{2} \quad Tinne Tuytelaars\textsuperscript{5} \quad Joost van de Weijer\textsuperscript{1,6}}  \\ 
\textsuperscript{1}Computer Vision Center, Barcelona, Spain \\
\textsuperscript{2}Media Integration and Communication Center, University of Florence, Italy \\
\textsuperscript{3}Bernoulli Institute, University of Groningen, the Netherlands \\ 
\textsuperscript{4}IDEAS Research Institute, Poland   \qquad
\textsuperscript{5}ESAT-PSI, KU Leuven, Belgium \\
\textsuperscript{6}Department of Computer Science, Universitat Autònoma de Barcelona, Spain \\
{\tt\small dgoswami@cvc.uab.es, simone.magistri@unifi.it}
}

\iclrfinalcopy % Uncomment for camera-ready version, but NOT for submission.
\begin{document}

\maketitle

\begin{abstract}
Vision-language models like CLIP are trained with the objective of aligning text and image pairs. Beyond text prompts alone, recent works show that exploiting few-shot image embeddings from a training set is effective for CLIP-based classification. In this work, we analyze mixing image and text prototypes from a bias-variance perspective and show that mixing prototypes acts like a variance shrinkage estimator. Naively mixing text and image prototypes combines two partially aligned spaces since the two modalities are not perfectly aligned. To address this, we project image prototypes onto the principal directions of the semantic text embedding space to obtain a task-semantic image subspace. Mixing the image prototypes with text embeddings in the task-semantic subspace improves few-shot classification. However, when the task-semantic subspace captures insufficient discriminative visual information, relying on this subspace alone can be suboptimal. On extensive experiments over several few-shot classification benchmarks, we show that combining a task-semantic mixed prototype classifier and an anisotropic image-specific classifier systematically outperforms existing methods.

\end{abstract}

\section{Introduction}

Vision-language models (VLMs) like CLIP~\citep{radford2021learning} are trained with a symmetric contrastive loss that maximizes the similarity between an image and its corresponding text while pushing it away from negative texts, and vice versa from text to image. This produces a shared embedding space in which semantically corresponding image and text representations are geometrically aligned. As a result, CLIP achieves strong zero-shot performance, where classification is performed using the cosine similarities between the test image representations and the text prompts encoded by the text encoder (text prototypes). However, the text and image embeddings in CLIP are not fully aligned and occupy different regions of the shared embedding space separated by a modality gap~\citep{liang2022mind,schrodi2025two,levi2025double}.

Multimodal Large Language Models (MLLMs)~\citep{liu2023visual} despite strong performance for generative tasks, struggle for discriminative tasks~\citep{zhang2024visually,feng2026generative}, making CLIP-based models more relevant for few-shot classification task. 
Several works explore the few-shot setting with CLIP in which only a few training images per class are available and show that the use of image embeddings in addition to text embeddings improves classification performance.
These include training-free approaches~\citep{zhang2021tip,wanghard,bendou2025proker} that combine image and text embeddings using different formulations, as well as methods that adapt the VLM by learning prompts~\citep{khattak2023maple,zhou2022cocoop} or adapters~\citep{silva2024closer,zanella2024cliplora}. 
In this paper, we analyze training-free mixed prototype-based classification and show that mixing shrinks the high variance of the few-shot image prototypes thereby improving image prototype estimation by controlling the bias-variance tradeoff. 

Although naive mixing is effective, the image and text representations are not perfectly aligned~\citep{liang2022mind}, resulting in the mixing of two misaligned prototypes. The image modality generally contains more detailed information such as background or instance-specific attributes such as object color which are not available in the text prompts, leading to the information imbalance between modalities~\citep{schrodi2025two}. 
We therefore ask the question: \textit{Can we use the class semantics to disentangle task-relevant image information from other visual variations for reliable few-shot classification?}

\begin{wrapfigure}{r}{0.57\textwidth}
  \centering
  \vspace{-15pt}
  \includegraphics[width=0.57\textwidth]{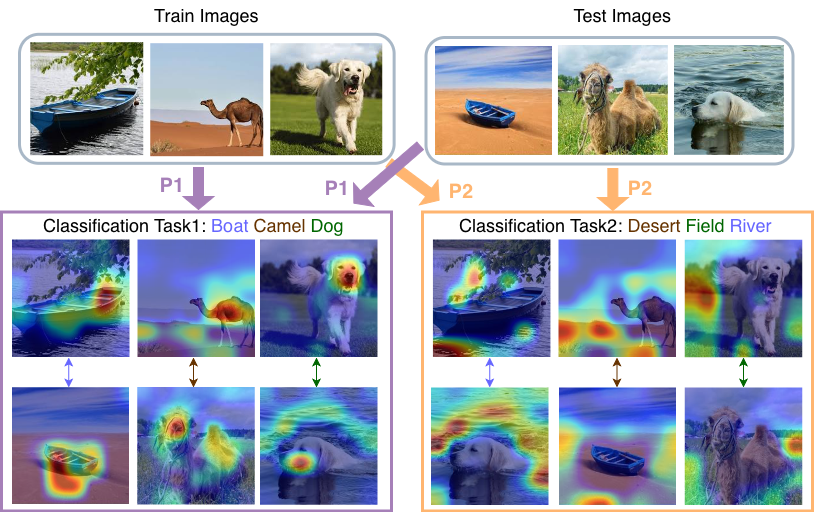}
  \vspace{-15pt}
  \caption{
  The latent space spanned by text prototypes defines a \textit{task-semantic subspace}.
  Task semantic regions denoted by attention maps are obtained by the proposed task-specific semantic projection P1 and P2. 
  }
  \label{fig:intro}
  \vspace{-10pt}
\end{wrapfigure}

We illustrate in~\cref{fig:intro} that the image space could be exploited differently based on the classification task defined by the class labels. 
We argue that the space spanned by the text prototypes defines a relevant task-semantic space. 
We exploit the cross-modal alignment in CLIP and decompose image prototypes into two components: a \textit{task-semantic subspace} to represent only the text label-specific information and a \textit{task-orthogonal subspace} to represent the remaining attributes and instance-specific details.
We show that mixing text and image prototypes in the task-semantic subspace improves classification performance over naively mixed prototype classifiers.

However, in out-of-distribution (OOD) datasets,
the text-based task-semantic subspace poorly reflects the task relevant structure of the image space due to the poor cross-modal alignment in CLIP.
To tackle this, we combine the mixed prototype classifier in the task-semantic subspace with a separate image classifier.
We summarize our contributions as follows:
\begin{itemize}
\item We analyze the mixing of image and text prototypes in CLIP from a bias–variance perspective for few-shot classification. We show theoretically and empirically (\cref{fig:mse}) that mixing acts as a shrinkage estimator, trading bias for variance reduction and yielding lower mean squared error than image-only prototypes.
\item We use the class text prototypes to define a task-semantic subspace and decompose image prototypes into task-semantic and orthogonal components. Our analysis (\cref{fig:task_semantic}) shows that the task-semantic subspace can concentrate discriminative visual information, while its effectiveness varies across tasks. Performing prototype mixing in this subspace improves prototype estimation and classification performance.
\item We complement the task-semantic classifier with an image-specific classifier that captures discriminative information beyond the task-semantic subspace. Across 15 diverse benchmarks, the proposed training-free method outperforms existing approaches from 1 to 16 shots, with particularly strong gains in the 1–4 shot regime, and generalizes across multiple contrastive vision-language backbones.
\end{itemize}

\section{Related Work}

\minisection{Vision-Language Models.}
VLMs learn joint image-text representations through large-scale contrastive pre-training. CLIP~\citep{radford2021learning} optimizes a symmetric contrastive loss over image-text pairs, producing a shared embedding space that enables strong zero-shot classification and other applications~\citep{baldrati2022effective,sain2023clip,yu2023convolutions,caselli2026spectralgcd}.
Despite this alignment objective,~\cite{liang2022mind} showed that image and text embeddings occupy distinct regions of the shared space separated by a \textit{modality gap}. 
Recent works have attempted to close or exploit this gap:~\cite{eslamimitigate} proposed methods to improve cross-modal alignment in CLIP, while~\cite{mistretta2025cross} exposed intra-modal misalignment via modality inversion. 
\cite{yu2024text} used text-guided attention to improve zero-shot robustness in CLIP.

\minisection{Few-Shot Classification.}
Among training-free methods, Tip-Adapter~\citep{zhang2021tip} constructs a key-value cache from few-shot image features and retrieves them at test time to augment zero-shot predictions. TIP-X~\citep{udandarao2023sus} extends this idea by populating the cache with support images retrieved or generated from class names. 
APE~\citep{zhu2023not} refines the cache by selecting only the most informative features via adaptive prior refinement. GDA~\citep{wanghard} proposed an ensemble of a zero-shot text classifier with an LDA image classifier. 
ProKeR~\citep{bendou2025proker} improves Tip-Adapter using a global regularizer in a reproducing kernel Hilbert space.

Training-based prompt learning methods~\citep{zhou2022coop,zhou2022cocoop} adapt CLIP by optimizing prompt tokens. 
MaPLe~\cite{khattak2023maple} extends this to multi-modal prompt learning across both encoders. Adapter-based methods tune lightweight modules: CLIP-Adapter~\citep{gao2024clip} adds residual feature adapters, MMA~\citep{yang2024mma} introduces multi-modal adapters, and CLIP-LoRA~\citep{zanella2024cliplora} applies low-rank adaptation to CLIP's encoders. 
In this work, we propose a training-free method that combines a task-semantic mixed prototype classifier with an image-based classifier. The proposed method can also be applied on top of training-based methods.

\section{Cross-Modal Prototype Mixing via Bias--Variance Analysis}
\label{sec:3}

A simple training-free method based on the CLIP vision encoder is the \textbf{Nearest Class Mean (NCM)} classifier, which estimates class prototypes as the empirical mean of training image features and assigns test samples to the nearest prototype. While effective, NCM performance depends on the quality of estimated visual prototypes. Here we analyze prototype estimation and introduce a \textbf{mixed prototype estimator} that improves estimation by controlling the bias--variance tradeoff.

\subsection{Image Prototype Estimation in the Few-Shot Regime}
Let $\mu_i^{*}$ be the true population image class mean for a given class, and $\hat{\mu}_i$ the empirical sample mean computed from $n$ samples used as the class prototype by the NCM classifier, with $\E[\hat{\mu}_i] = \mu_i^{*}$, $\Cov[\hat{\mu}_i] = \frac{\Sigma_i^*}{n}$, where $\Sigma_i^*$ is the population covariance of the class. Recall that, for a generic estimator $\hat{\theta}$ of a parameter $\theta^*$, the Mean Squared Error (MSE) is defined as:
\begin{equation}
    \mathrm{MSE}(\hat{\theta}, \theta^*) 
    = \mathrm{Bias}^2(\hat{\theta}, \theta^*) 
    + \mathrm{Var}(\hat{\theta}).
\end{equation}
The sample mean is unbiased, thus $\mathrm{Bias}(\hat{\mu}_{i}, \mu_i^*) = \E[\hat{\mu}_i] - \mu_i^* = 0$. Therefore the MSE of the image mean estimator reduces to the variance term:  
\begin{equation}
\label{eq:MSE_ncm}
    \mathrm{MSE}(\hat{\mu}_{i}, \mu_i^{*})= \left\|
         \E[\hat{\mu}_i] - \mu_i^{*}
      \right\|^2
      + \operatorname{tr}\!\left(
            \Cov[\hat{\mu}_i]
        \right)
    = \frac{1}{n} \operatorname{tr}(\Sigma_i^*).
\end{equation}
This result shows that the estimation error of the class prototype scales inversely with the number of samples $n$. This classical decomposition explains why the NCM classifier becomes unreliable in the few-shot regime: the variance of the image mean estimator becomes large, leading to unreliable prototype estimates and degraded classification performance. CLIP also has a text encoder which, when fed with a text prompt, produces text prototypes for zero-shot prediction. 
We now present a better estimator of the image population mean that exploits these text prototypes.

\subsection{Mixed Image and Text Prototype Estimator}
Let $\mu_t^{*}$ be the text prototype for class $c$, obtained deterministically by feeding a fixed prompt template to text encoder, so that $\E[\mu_t^{*}] = \mu_t^* \text{ and } \Cov [\mu_t^{*}] = 0$. We define the \textit{mixed prototype estimator} as:
\begin{equation}
\label{eq:mixing}
    \hat{\mu}_{\text{mix}} = \lambda \hat{\mu}_i + (1-\lambda) \mu_t^*,
\end{equation}
where $\lambda \in [0,1] $ is the mixing coefficient controlling the contribution of the image prototype. We propose to use this mixed prototype for each class during classification, since it provides a better estimate of the population class mean than the standard image prototype. Classification is then performed using NCM with the mixed prototype $\hat{\mu}_{\text{mix}}$.

The MSE of the mixed prototype estimator with respect to the true population image mean can be decomposed into the sum of the squared bias and variance (see~\cref{app:mse-derivations} for the full derivation):
\begin{equation}
\label{eq:mse-mix}
\begin{aligned}
    \text{MSE}(\hat{\mu}_{\text{mix}}, \mu_i^{*})
    =& (1-\lambda)^2 || \mu_t^* -\mu_i^* ||^2 + \lambda^2 \frac{1}{n} \text{tr} (\Sigma_i^*).
\end{aligned}
\end{equation}

\minisection{Mixing with text prototype acts as a variance shrinkage operator.} Eq.~\ref{eq:mse-mix} reveals that prototype mixing behaves as a shrinkage estimator. While mixing introduces a bias term determined by the modality gap $|| \mu_t^* -\mu_i^* ||^2$ between population image mean and text, at the same time it reduces the variance of the estimator through the $\lambda^2$ factor scaling the image covariance trace. 
When the modality gap is moderate compared to the variance term, mixing can reduce overall MSE by trading a small increase in bias for a larger reduction of variance.

\begin{wrapfigure}{r}{0.51\textwidth} % {r} for right, {l} for left
  \centering
  \vspace{-3pt}
  \includegraphics[width=0.51\textwidth]{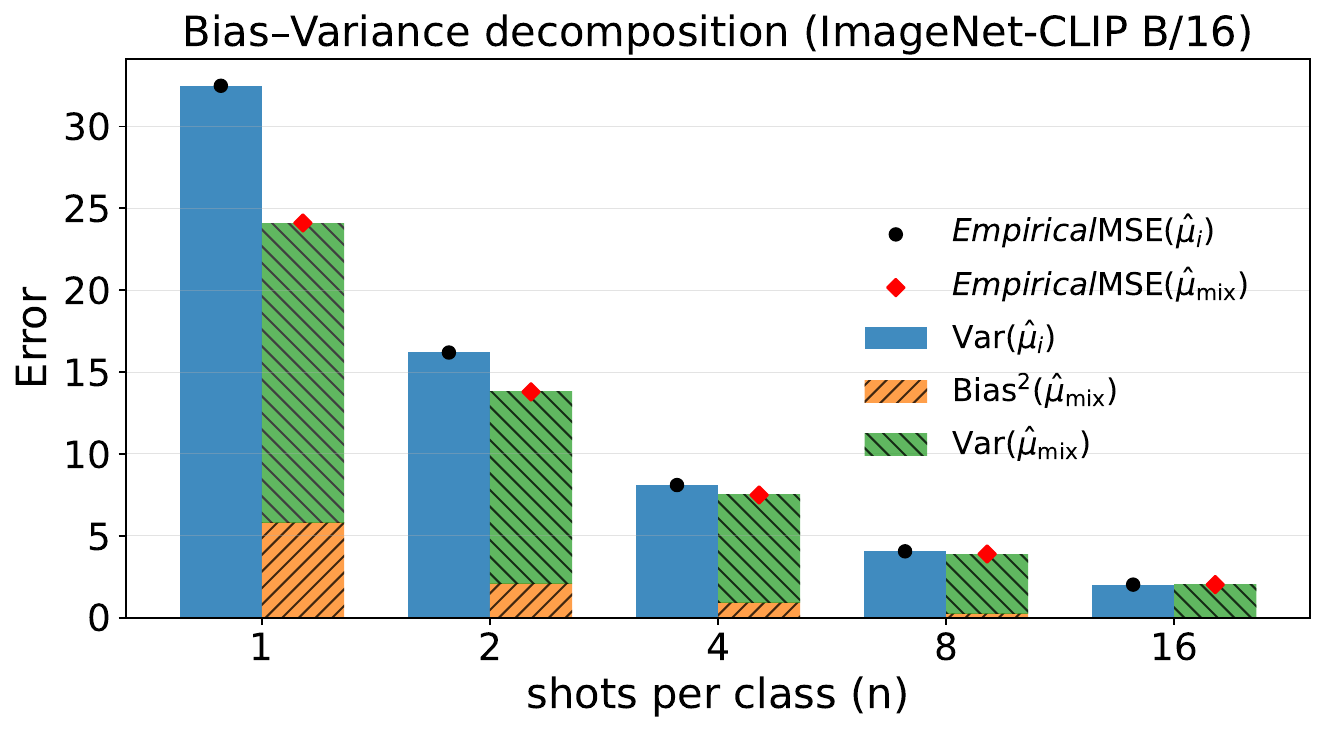}
  \vspace{-18pt}
  \caption{MSE of image and mixed prototypes as a function of the number of shots using CLIP-B/16.}
  \label{fig:mse}
  \vspace{-10pt}
\end{wrapfigure}

In \cref{fig:mse}, we analyze this decomposition on ImageNet.
We approximate the population statistics $\mu_i^*$ and $\Sigma_i^*$ using all training samples per class. For the text prototype $\mu_t^{*}$, we use the standard CLIP prompt template “a photo of a \{class name\}”. We then simulate the few-shot setting by repeatedly sampling $n$ examples per class with replacement for $n \in \{1,2,4,8,16\}$. For each shot number $n$, we compute the empirical MSE of the image prototype $\hat{\mu}_{i}(n)$ and the mixed prototype $\hat{\mu}_{\text{mix}}(n)$ with respect to the estimated population mean $\mu_i^*$. The results are averaged across multiple Monte Carlo trials and classes. 
The empirical MSE values are shown as dots in \cref{fig:mse}, while the bars illustrate the bias and variance components from the theoretical decomposition. 
The mixing parameter $\lambda$ is selected from a fixed grid for each shot level to minimize the empirical MSE.
In the lower shots, image prototypes deviate significantly from the population means, resulting in higher MSE. Using mixed prototypes (green+orange) reduces the MSE, providing more accurate estimates of the population mean and a better bias–variance trade-off than the image prototype estimator.

\minisection{Limitations of naive prototype mixing.} 
While mixed prototypes reduce the estimation error of the class mean, there is a limitation:
mixing is performed in the full embedding space, implicitly combining components of the image representation that may not be semantically related to the text prototype. 
In CLIP, image and text representations are only partially aligned~\citep{liang2022mind}. As a result, certain directions in the image embedding space correspond to semantic information captured by the text encoder, while others encode image-specific variations~\citep{schrodi2025two}. 
Since text prototypes provide no information about image-specific directions, we propose mixing them only with image components in the text subspace to improve classification.

\section{Task-Semantic Projection}
\label{sec:4}

CLIP's contrastive training encourages image and text representations to align. Consequently, image class means may be approximately aligned with their corresponding text prototypes, although they can also contain information absent in the text descriptions. At inference, the text prototypes of the $C$ candidate classes define the classification task. We use them to identify the \emph{task-semantic subspace}, allowing to isolate the image mean components along text-defined directions.

Let $T = [\mu_{t,1}^*, \ldots, \mu_{t,C}^*] \in \mathbb{R}^{d \times C}$ be the matrix of all the text prototypes.
We compute a truncated Singular Value Decomposition (SVD) of $T$ at rank $k$, $T_k = U_{k}\Sigma_{k} V_{k}^{\top} \in \mathbb{R}^{d \times C}$, where $U_k \in \mathbb{R}^{d \times k}$, $\Sigma_k \in \mathbb{R}^{k \times k}$, $V_k \in \mathbb{R}^{C \times k}$ contain the top-$k$ left singular vectors,
singular values, and right singular vectors,
respectively. We define the matrix
% $P = U_k U_k^{\top} \in \mathbb{R}^{d \times d}$
\begin{equation}
    P = U_k U_k^{\top} \in \mathbb{R}^{d \times d}
\end{equation}
as the \textit{task-semantic projector} onto the $k$-dimensional
principal subspace spanned by left singular vectors of the text prototype matrix. The range $U_k$ forms the \textit{task-semantic subspace}.

For any image prototype $\mu_i \in \mathbb{R}^d$
(for brevity, we write $\mu_i = \mu_{i,c}$),
we can decompose it as:
\begin{equation}
\label{eq:decomposition-1}
\mu_i
=
P \mu_i
+
(I_d - P)\mu_i = \mupar  + \muort ,
\end{equation}
where $\mupar=P\mu_i$ is the \textit{task-semantic component} of the image prototype, which is its projection onto the subspace defined by the class text prototypes. The residual $\muort=(I_d -P) \mu_i$ is the task-orthogonal complement, where $I_d \in \mathbb{R}^{d \times d}$ is the identity matrix. Intuitively, $\mupar$ retains image information along the dominant text directions for the given classification task,  while $\muort$ captures additional information outside those directions that may still aid classification.

\newcolumntype{C}[1]{>{\centering\arraybackslash}p{#1}}

\begin{figure}[t]

\centering
% ---- RIGHT: attention map table ----
\begin{minipage}[t]{0.42\textwidth}
\vspace{0pt}
    \centering
    \setlength{\tabcolsep}{2pt}
    \renewcommand{\arraystretch}{0.4}
    \begin{tabular}{c C{0.24\textwidth} C{0.24\textwidth}  C{0.24\textwidth}}
        & \scriptsize\textbf{Aircrafts} & \scriptsize\textbf{Caltech101} & \scriptsize\textbf{UCF101} \\

        \adjustbox{valign=c,angle=90}{\scriptsize\textbf{Orthogonal}} &
        \includegraphics[width=0.92\linewidth]{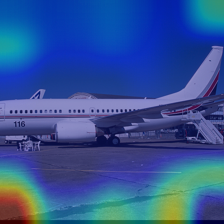} &
        \includegraphics[width=0.92\linewidth]{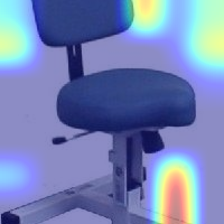} &
        \includegraphics[width=0.92\linewidth]{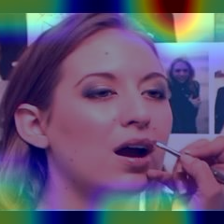} \\

        \adjustbox{valign=c,angle=90}{\scriptsize\textbf{Semantic}} &
        \includegraphics[width=0.92\linewidth]{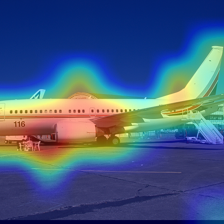} &
        \includegraphics[width=0.92\linewidth]{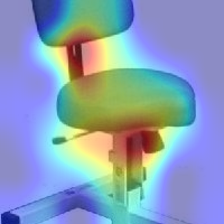} &
        \includegraphics[width=0.92\linewidth]{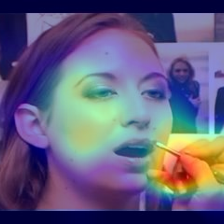} \\

        & \tiny ``a photo of a \textit{ATR-72}, a type of aircraft.''
        & \tiny ``a photo of a \textit{chair}.''
        & \tiny ``a photo of a person doing \textit{Apply Lipstick}." \\
    \end{tabular}
\end{minipage}
\hfill
% ---- LEFT: analysis plot ----
\begin{minipage}[t]{0.57\textwidth}
\vspace{0pt}
    \centering
    \includegraphics[width=\linewidth]{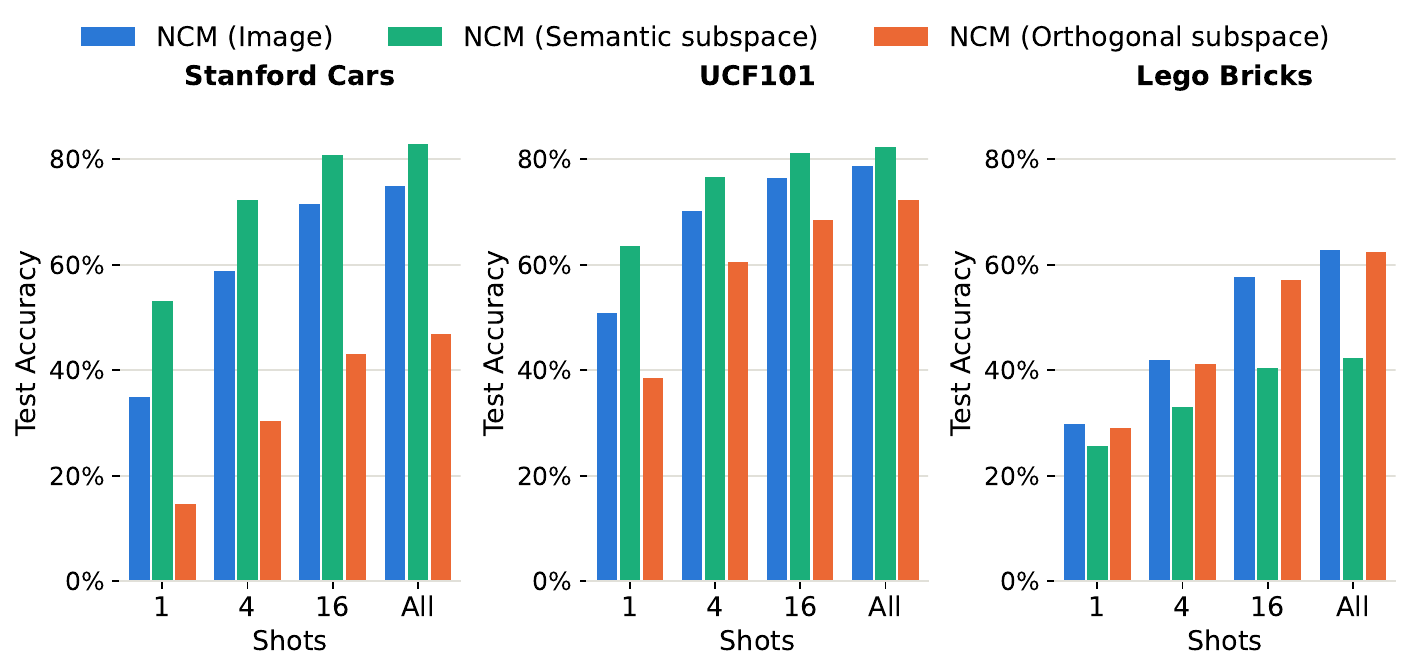}
\end{minipage}

\vspace{-5pt}
\caption{Impact of the \textit{task-semantic} projection: (left) Attention maps reveal that visual features projected on the task-semantic subspace focus on the text information; (right) Using NCM classifier for standard datasets (Stanford Cars~\citep{krause20133d} and UCF101~\citep{soomro2012ucf101}) on CLIP-ViT-B/16, we show that task-semantic subspace is more discriminative than the full image space while the task-orthogonal subspace performs poorly. For OOD datasets like Lego Bricks~\citep{kaggle_lego}, the full image space and the orthogonal subspace is more discriminative than the semantic subspace due to poor cross-modal alignment in CLIP.}
\vspace{-15pt}
\label{fig:task_semantic}
\end{figure}

To visualize the image regions associated with the task-semantic subspace and its orthogonal complement, we use GradCAM~\citep{Selvaraju2020}. Using the text prompts for all classes in a dataset, we compute the projector $P$. For an image feature $f_i$, we evaluate the energy $f_i P f_i^{\top}$ and backpropagate this score to the last visual layer, highlighting regions that contribute to the task-semantic subspace. In contrast, using the orthogonal projection $(I-P)$ reveals complementary information. As shown in~\cref{fig:task_semantic} (left), the semantic component focuses on regions identified by the text space, while the orthogonal component captures instance-specific details. 
We demonstrate in~\cref{fig:task_semantic} (right) that the accuracy of an NCM classifier after projecting image prototypes onto the task-semantic subspace improves significantly over naive NCM on Stanford Cars and UCF101. This observation holds irrespective of the number of shots, suggesting a better class separation than the original image prototypes.

Although the proposed semantic projection works well for standard in-distribution datasets, we show in~\cref{fig:task_semantic} (right) that the task-semantic subspace performs poorly for OOD datasets like Lego Bricks where the original image space is more discriminative.
We also observe that the task-orthogonal subspace is much less discriminative for Stanford Cars and UCF, but is more discriminative for Lego Bricks.
These results suggest that the quality of the semantic subspace varies across tasks or datasets and depends on the alignment between the image and the text embedding space.
We now discuss how to measure the existing cross-modal alignment in CLIP.
 
\minisection{Analysis of Cross-Modal Alignment.}
The principal angles between the image and text prototype spaces
measure cross-modal alignment, indicating how well the semantic structure in text matches the geometry of the image space. Computing SVD of text and image prototypes $T$ and $I$, we obtain the cross-modal matrix $M$ as:
\begin{equation}
\label{eq:cross-modal}
     T = U_{t}\Sigma_{t} V_{t}^{\top} \in \mathbb{R}^{d \times C}, \enspace
     I = U_{i}\Sigma_{i} V_{i}^{\top} \in \mathbb{R}^{d \times C}, 
     \quad
      M = U_{t}^{\top}U_{i} = U_{M}S_MV_{M}^{\top} \in \mathbb{R}^{C \times C}
\end{equation}
where the singular values in $S_M$ correspond to the \textit{cosines of  the principal angles}.

\begin{wrapfigure}{r}{0.45\textwidth}
  \centering
  \vspace{-2pt}
  \includegraphics[width=\linewidth]{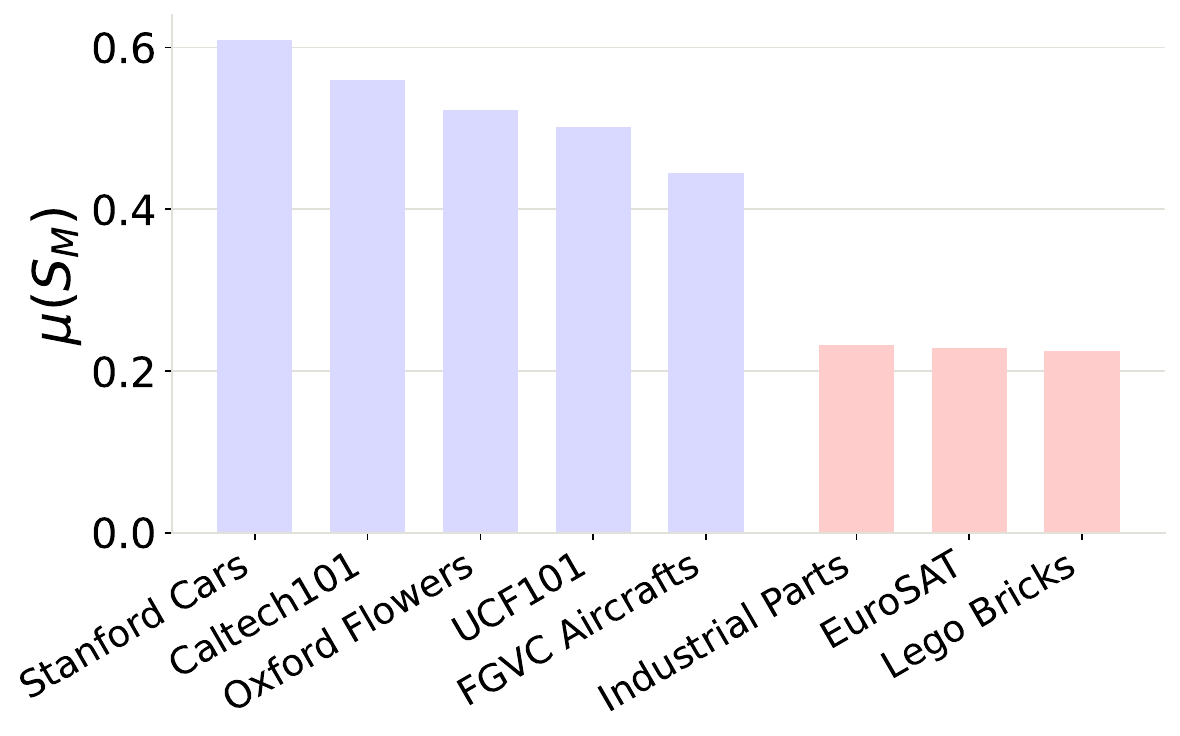}
  \vspace{-15pt}
  \caption{Analysis of $\mu(S_M)$.}
  \vspace{-10pt}
  \label{fig:principal-angles}
\end{wrapfigure}

The mean of the principal angle cosines $\mu(S_M)$, shown in~\cref{fig:principal-angles} for 16-shots, indicates that CLIP-ViT-B/16 exhibits weak alignment for some datasets like EuroSAT, Lego Bricks and stronger alignment for others like Stanford Cars, UCF101. Refer to~\cref{app:datasets} for analysis on more datasets.
The poor alignment could be due to poor semantic quality of the class labels (e.g. \texttt{Plate 1$\times$2}, \texttt{Brick 1$\times$3} in Lego Bricks) or could be due to a strong domain shift from training data distribution (e.g. satellite images in EuroSAT~\citep{helber2019eurosat}).

Due to poor alignment in these datasets, the proposed semantic projector does not manage to extract relevant discriminative features from the image space resulting in a weaker task-semantic subspace, and hence poor NCM performance.
In such cases, the full image space and the orthogonal component have more discriminative information.

Furthermore, even with good cross-modal alignment, the task-orthogonal component may contain discriminative information absent from text~\citep{schrodi2025two}.
We therefore separate these two sources of information: mix text and semantically projected image prototypes
while handling image-specific additional information separately through a classifier in the full image space.

\section{Proposed Training-free Classifier for Few-Shot Classification}
\label{sec:5}

In this section, we propose a training-free classifier for few-shot settings that combines a classifier in the task-semantic subspace with an image-based classifier. 

\minisection{Mixing Prototypes in the Task Semantic Subspace.} The task-semantic projection improves classification by retaining more discriminative information, but the projected few-shot means $\hat{\mu}_i^{||} = P \hat{\mu}_i$ remains noisy. As discussed in Section~\ref{sec:3}, mixing with a text prototype reduces image mean estimation variance. Unlike mixing in the full image space, the relevant estimation target is now the projected population mean $\mu_i^{||,*} = P\mu_i^*$. We therefore mix prototypes in this subspace and use the resulting \textbf{Proj+Mix} prototypes to construct an NCM classifier in the task-semantic subspace (TSS), with logits computed using cosine similarity.
We compute the logits s$_{TSS}$
as follows:
\begin{equation}
\label{eq:tamp}
\begin{aligned}
\mu_{\mathrm{Proj+Mix},c}
&= \lambda \mu_{i,c}^{\parallel}+(1-\lambda)\mu_{t,c}^*,   \qquad
\text{s}_{\mathrm{TSS},c}
= \frac{Pf_{\mathrm{test}}}{\lVert P f_{\mathrm{test}} \rVert}
\cdot
\frac{\mu_{\mathrm{Proj+Mix},c}}
{\lVert \mu_{\mathrm{Proj+Mix},c} \rVert},
\enspace c=1,\ldots,C,
\end{aligned}
\end{equation}
where $\mu_{i,c}^{\parallel}=P\mu_{i,c}$ is the projected image prototype of class $c$, $Pf_{\text{test}}$ is the projected test image feature and $\lambda$ controls the mixing weight.
Note that we apply the task-semantic projection to the test image features to classify them using the NCM classifier with Proj+Mix estimated prototypes. In~\cref{app:mixing-analysis}, we analyze why mixing prototypes in the full image space is suboptimal.

To incorporate additional information beyond the text-based task semantics, we propose to use a Mahalanobis image-based classifier. 

\minisection{Image-based Classifier.}
We compute the global covariance $\Sigma$ using the features centered on the class means. We use a shrinkage regularized form of the covariance as $A = (N-1)\Sigma + \mathrm{tr}(\Sigma) I$ where $N$ is the total number of samples. 
Now, we compute $A_M = A^{-1/2}$ which will be used to compute the Mahalanobis cosine similarity between the test features $f_{\text{test}}$ and the prototypes $\mu_{i,c}$. 
We compute the logits s$_{IM}$ as follows:
\begin{equation}
    \label{eq:image-classifier}
    \text{s}_{IM,c} = \frac{A_M f_{\text{test}}}{\lVert A_M f_{\text{test}} \rVert} \cdot\frac{A_M \mu_{i,c}}{\lVert A_M \mu_{i,c} \rVert}, \qquad c=1,\ldots,C,
\end{equation}
where $A_M$ is the Mahalanobis projection in the image space. 
By exploiting the global image covariance, this classifier captures the anisotropy of the image space.

\minisection{Cross-Modal Prototype (CMP) Classifier.} The final classifier combines the logits from task-semantic subspace classifier s$_{TSS}$ and the image-based classifier s$_{IM}$. The output logits s$_{CMP}$ are calculated as:
\begin{equation}
    \label{eq:ensemble}
    \text{s}_{CMP} = \text{s}_{TSS} + \gamma \enspace \text{s}_{IM}
\end{equation}
where $\gamma$ is a hyperparameter combining the two classifiers.
This final classifier combines the strengths of the Proj+Mix estimator, exploiting the text-based semantics and reducing the variance in the lower shots with an anisotropic image classifier that can exploit reliable covariance estimates in the higher-shots.

While $\lambda$ and $\gamma$ could be tuned on a validation set following~\cite{zhang2021tip,zhu2023not,wanghard}, we also propose a \textit{validation-free} approach to select $\lambda$ and $\gamma$. 
Intuitively, the mixing coefficient $\lambda$ used in~\eqref{eq:tamp} for the task-semantic subspace classifier could be selected based on shots. Lower shots imply lower weight $\lambda$ for image prototypes due to higher variance (\eqref{eq:MSE_ncm}).
We propose using $\lambda$: \{1: 0.1,  2: 0.2, 4: 0.3, 8: 0.4, 16: 0.5\} for respective shots.

To select $\gamma$ in~\eqref{eq:ensemble}, we propose to exploit the existing cross-modal alignment as follows: 
\begin{equation}
    \label{eq:gamma}
    \gamma = (1 - \mu(S_M))^2 * n/(n+1)
\end{equation}
where $\mu(S_M)$ refer to the mean of principal angle cosines $S_M$ (from~\eqref{eq:cross-modal}) and $n$ refers to the number of shots. Here, $\gamma$ gives us a squared cross-modal misalignment score scaled by a few-shot reliability term $n/(n+1)$ since principal angle cosines are computed from few-shot training samples.

\section{Experimental Results}

\vspace{-5pt}
\minisection{Datasets.} We perform evaluation on 15 datasets comprising of standard in-distribution and out-of-distribution (OOD) datasets. 
Standard datasets include ImageNet (ImgNet), Caltech101 (CAL), SUN397 (SUN), UCF101 (UCF), FGVC Aircrafts (AIR), Stanford Cars (CARS), Oxford Pets (PETS), Flowers102 (FLWR), Food101 (FOOD) and Flying Insects (INSCT). OOD datasets include a satellite image dataset - EuroSAT (ESAT), a texture dataset - Describable Textures (DTD), a yoga pose dataset (YOGA), a lego bricks dataset (LEGO) and an industrial parts dataset (INDS). We provide more details in~\cref{app:datasets}.

\minisection{Implementation Details.} We select 1, 2, 4, 8, or 16 random samples per class to evaluate few-shot settings. To ensure fair comparison with existing training-free methods, we reproduce the baselines in the same evaluation settings and use three variants of CLIP - ViT-B/16, Siglipv2-B/16 and the recently proposed PE-Core-B/16~\citep{bolya2026perception}. For all experiments in~\cref{tab:table1}, the hyper-parameters of all methods
are selected using the validation set following standard practice~\citep{zhang2021tip,zhu2023not,wanghard}. We select $\lambda$ and $\gamma$ for our method (CMP) via grid search over 10 values in the range [0, 1]. 
We also present the \textit{validation-free} setting performance in~\cref{tab:table2}. For the decomposition of image prototypes, we select the top-$k$ components based on the explained variance ratio which covers 99\% of the variance. 
Following existing methods, we apply 10 augmentations per image.
All experiments report average accuracy over 3 random seeds.

\begin{table*}[t!]
    \def\arraystretch{1.1}
    \centering
    % \footnotesize
    \caption{Few-shot classification performance (mean over 3 seeds) for training-free methods on 15 datasets using CLIP-ViT-B/16. Best results in each group are in \textbf{bold} and second best \underline{underlined}.}
    \vspace{-5pt}
    \begin{adjustbox}{max width=\textwidth}
    \begin{tabular}{llccccccccccccccc|c}
    \toprule
    & & \multicolumn{10}{c}{\cellcolor{blue!15}Standard Datasets} & \multicolumn{5}{c}{\cellcolor{red!20}OOD Datasets} & \\
\cmidrule(lr){3-12} \cmidrule(lr){13-17}
     Shot & Method & SUN & CARS & FOOD & CAL & FLWR & ImgNet & UCF & INSCT & AIR & PETS & YOGA & DTD & LEGO & ESAT  & INDS  & Avg \\
    \midrule

& \emph{Zero-Shot} & 62.6 & 65.3 & 86.1 & 92.9 & 71.4 & 68.8 & 66.8 & 35.4 & 24.8 & 89.1 & 6.6 & 44.4 & 8.9 & 47.7 & 27.2 & 53.2 \\

\cmidrule(lr){1-18}
\cellcolor{gray!0} \multirow{4}{*}{\texttt{1}} 
& TIP-Adp.  & 65.1 & 67.3 & 85.9 & 93.6 & 82.8 & 69.2 & 71.3 & 41.8 & 28.4 & 89.4 & 7.8 & 51.0 & 30.7 & 63.9 & 51.4 & 60.0 \\

& APE & \textbf{66.7} & \underline{68.6} & \underline{86.1} & \textbf{94.2} & \textbf{87.5} & \textbf{69.6} & \underline{72.4} & 45.2 & \underline{29.0} & \underline{90.9} & \underline{9.3} & \textbf{53.9} & \underline{33.7} & 66.8 & 57.2 & \underline{62.1} \\

& GDA  & 65.6 & 65.4 & \textbf{86.2} & 93.0 & 80.9 & 69.1 & 72.3 & 47.0 & 26.7 & 89.3 & 7.4 & 48.1 & 31.3 & \underline{68.9} & \underline{59.7} & 60.7 \\

& ProKeR & 63.8 & 66.4 & 86.0 & 93.6 & 84.3 & \underline{69.3} & 70.7 & \underline{47.6} & 27.6 & 89.6 & \underline{9.3} & 49.7 & 33.2 & 61.8 & 58.7 & 60.8 \\

\hlrow \cellcolor{white}{}  & CMP & \underline{66.6} & \textbf{69.9} & 85.9 & \underline{93.7} & \underline{87.4} & 68.9 & \textbf{73.1} & \textbf{57.3} & \textbf{30.9} & \textbf{91.0} & \textbf{11.2} & \underline{53.7} & \textbf{34.3} & \textbf{72.4} & \textbf{65.3} & \textbf{64.1} \\

\cmidrule(lr){1-18}

\cellcolor{gray!0} \multirow{4}{*}{\texttt{2}} 

& TIP-Adp.  & 67.3 & 68.3 & 86.1 & 94.0 & 87.0 & 69.4 & 73.4 & 46.9 & 29.1 & 90.1 & 8.9 & 54.2 & 32.5 & 68.8 & 59.1 & 62.3 \\

& APE   & \underline{69.1} & 69.8 & 86.1 & \underline{94.5} & 90.8 & \textbf{70.0} & \textbf{77.1} & 49.9 & 30.1 & \underline{91.6} & 11.1 & \underline{58.2} & 38.1 & 72.5 & 66.4 & 65.0 \\

& GDA  & 68.6 & \underline{70.6} & \textbf{86.3} & 93.8 & 89.2 & 69.5 & 75.8 & \underline{63.5} & \underline{31.0} & 90.5 & \underline{12.6} & 54.5 & \underline{49.1} & \underline{75.0} & \underline{74.8} & \underline{67.0} \\

& ProKeR & 67.4 & 67.8 & \underline{86.2} & 93.8 & \underline{91.1} & \underline{69.9} & 74.4 & 60.3 & 29.8 & 89.8 & \underline{12.6} & 55.4 & 44.3 & 72.8 & 72.5 & 65.9 \\

\hlrow  \cellcolor{white}{}  & CMP & \textbf{70.0} & \textbf{73.5} & \underline{86.2} & \textbf{94.6} & \textbf{92.7} & 69.7 & \underline{76.5} & \textbf{69.0} & \textbf{34.0} & \textbf{92.4} & \textbf{16.3} & \textbf{59.8} & \textbf{51.3} & \textbf{77.3} & \textbf{76.6} & \textbf{69.3} \\

\cmidrule(lr){1-18}

\cellcolor{gray!0} \multirow{4}{*}{\texttt{4}} 

& TIP-Adp.  & 68.8 & 70.2 & 86.1 & 94.2 & 89.2 & 69.6 & 73.1 & 50.5 & 31.4 & 90.0 & 9.3 & 56.6 & 38.5 & 73.4 & 69.5 & 64.7 \\

& APE   & 70.8 & 72.7 & 86.4 & \underline{94.8} & 92.5 & 70.4 & 76.5 & 56.0 & 33.5 & \underline{91.6} & 14.1 & \underline{61.4} & 45.9 & 76.6 & 76.5 & 68.0 \\

& GDA  & \underline{72.1} & \underline{74.6} & \textbf{86.6} & 94.6 & 94.3 & \underline{70.6} & \underline{76.9} & \underline{74.6} & \underline{36.9} & 91.3 & \underline{20.0} & \underline{61.4} & \underline{64.5} & \underline{80.3} & \underline{86.0} & \underline{72.3} \\

& ProKeR & 70.1 & 72.0 & 86.3 & 94.3 & \underline{94.4} & \textbf{70.9} & 76.4 & 70.0 & 35.2 & 90.1 & 19.3 & 61.2 & 54.9 & 79.4 & 84.0 & 70.6 \\

\hlrow  \cellcolor{white}{}  & CMP & \textbf{72.8} & \textbf{77.2} & \underline{86.5} & \textbf{95.2} & \textbf{95.3} & 70.5 & \textbf{78.6} & \textbf{76.8} & \textbf{39.0} & \textbf{92.2} & \textbf{25.2} & \textbf{64.3} & \textbf{65.3} & \textbf{82.1} & \textbf{87.1} & \textbf{73.9} \\

\cmidrule(lr){1-18}

\cellcolor{gray!0} \multirow{4}{*}{\texttt{16}} 

& TIP-Adp.  & 71.5 & 74.0 & 86.4 & 94.5 & 92.9 & 70.7 & 76.0 & 59.5 & 36.7 & 90.0 & 12.7 & 62.2 & 51.8 & 77.3 & 72.5 & 68.6 \\

& APE   & 73.4 & 77.5 & 86.6 & 95.0 & 94.9 & 71.6 & 78.6 & 65.3 & 40.6 & 92.1 & 20.8 & 68.4 & 61.2 & 80.9 & 80.9 & 72.5 \\

& GDA  & 75.9 & \underline{83.1} & \underline{87.3} & 95.7 & 97.6 & 72.2 & 83.4 & \underline{85.6} & \underline{49.8} & \underline{93.6} & \underline{39.7} & 70.9 & \underline{85.3} & \underline{89.2} & \underline{93.0} & \underline{80.2} \\

& ProKeR & \underline{76.1} & 82.9 & 86.5 & \underline{96.0} & \textbf{98.4} & \textbf{73.3} & \textbf{84.2} & 85.5 & 49.3 & 90.5 & 36.0 & \underline{71.7} & 77.4 & 88.5 & 92.1 & 79.2 \\

\hlrow \cellcolor{white}{} &  CMP & \textbf{76.7} & \textbf{83.6} & \textbf{87.4} & \textbf{96.2} & \underline{98.1} & \underline{72.5} & \underline{83.8} & \textbf{87.7} & \textbf{50.6} & \textbf{94.0} & \textbf{41.8} & \textbf{72.7} & \textbf{85.4} & \textbf{89.8} & \textbf{93.5} & \textbf{80.9} \\

\bottomrule
    \end{tabular}
    \end{adjustbox}
\label{tab:table1}
\vspace{-5pt}
\end{table*}

\minisection{Baselines.} We compare our method with several training-free and training-based approaches. Among training-free methods, we use Zero-Shot CLIP~\citep{radford2021learning}, 
Tip-Adapter~\citep{zhang2021tip}, APE~\citep{zhu2023not}, GDA~\citep{wanghard} and ProKeR~\citep{bendou2025proker}. 
We also evaluate our approach on top of the adapted models from MaPLe~\citep{khattak2023maple} and CLIP-LoRA~\citep{zanella2024cliplora}.

\subsection{Comparison with Training-free Methods} 
\cref{tab:table1} reports the comparison with training-free approaches across 15 datasets. The results show that the proposed method CMP outperforms existing approaches across all settings with more significant improvements on average in 1-Shot (+2\% over APE), 2-shot (+2.3\% over GDA) and 4-Shot (+1.6\% over GDA) settings. We also improve over GDA in 16-Shot settings by 0.5\% on average. Existing methods fare well either on lower- or higher-Shot settings (APE does better than GDA in 1-Shot while GDA is much better than APE in 2-, 4- and 16-Shot) while CMP outperforms them in most cases on all shots across standard and OOD datasets. 

\cref{tab:table2} presents the performance comparison across different models and shots for 14 datasets in the \textit{validation-free} setting where we consider no access to dataset-specific validation sets for our approach. While all existing methods need to tune the hyperparameters on a reference dataset like ImageNet as shown in~\cite{bendou2025proker}, our method CMP works without any reference validation set. As proposed in~\cref{sec:5}, we select $\lambda$ based on shots and $\gamma$ using the cross-modal misalignment score (\eqref{eq:gamma}). CMP outperforms all methods consistently using 3 VLMs on average over all shots (+1.8\% over GDA on CLIP-ViT-B/16, +1.3\% over ProKeR on SigLIPv2-B/16 and +1.6\% over GDA on PE-Core-B/16). This further demonstrates the robustness of our method with reduced reliance on availability of validation sets. We provide more detailed results across all datasets on CLIP-ViT-B/16 in Appendix~\cref{app:valfree}.

\begin{table*}[t!]
    \def\arraystretch{1.1}
    \centering
    % \footnotesize
    \caption{Few-shot classification accuracy (mean over 3 seeds) on 14 datasets using different models in the \textit{validation-free setting} (Validation on ImageNet for existing methods while ours do not require any validation set). Best results in each group are in \textbf{bold} and second best \underline{underlined}.}
    \vspace{-5pt}
    \begin{adjustbox}{max width=\textwidth}
    \begin{tabular}{lcccccc|cccccc|cccccc}
    \toprule
 & \multicolumn{6}{c}{\cellcolor{blue!10}CLIP-ViT-B/16} & \multicolumn{6}{c}{\cellcolor{red!15}SigLIPv2-B/16} & \multicolumn{6}{c}{\cellcolor{green!15}PE-Core-B/16} \\
\cmidrule(lr){2-7} \cmidrule(lr){8-13} \cmidrule(lr){14-19}
     Method & 1 & 2 & 4 & 8 & 16 & Avg & 1 & 2 & 4 & 8 & 16 & Avg  & 1 & 2 & 4 & 8 & 16 & Avg \\
    \midrule

TIP-Adp.  & 53.9 & 55.7 & 57.1 & 58.0 & 66.9 & 58.3 & 70.1 & 71.2 & 72.6 & 75.5 & 79.3 & 73.7 &  72.2 & 70.8 & 76.1 & 72.8 & 78.4 & 74.1 \\

APE & 57.3 & 57.5 & 61.3 & 64.1 & 69.8 & 62.0 & 70.6 & 72.7 & 74.2 & 77.4 & 81.9 & 75.3 & 70.0 & 73.3 & 74.9 & 75.5 & 80.3 & 74.8 \\

GDA  & 58.3 & \underline{65.6} & \underline{71.3} & \underline{76.5} & \underline{80.3} & \underline{70.4} & 62.0 & 75.0 & \underline{81.9} & \textbf{85.6} & \underline{86.9} & 78.3 & 72.8 & \underline{76.2} & \underline{80.6} & \underline{84.4} & \underline{86.3} & \underline{80.1} \\

ProKeR & \underline{59.7} & 63.2 & 66.9 & 73.4 & 79.1 & 68.5 & \underline{74.6} & \underline{77.1} & 80.5 & 83.3 & \textbf{87.1} & \underline{80.5} & \underline{74.0} & 76.1 & 78.5 & 82.7 & 86.0 & 79.4 \\

\hlrow \cellcolor{white}{CMP}  & \textbf{61.7} & \textbf{68.1} & \textbf{73.2} & \textbf{77.3} & \textbf{80.4} & \textbf{72.2} & \textbf{74.7} & \textbf{79.0} & \textbf{82.9} & \underline{85.5} & 86.8 & \textbf{81.8} & \textbf{74.3} & \textbf{78.3} & \textbf{82.6} & \textbf{85.8} & \textbf{87.7} & \textbf{81.7} \\

\bottomrule
    \end{tabular}
    \end{adjustbox}
\label{tab:table2}
% \vspace{-5pt}
\end{table*}

% In the document
\begin{figure}[t]
\centering

% -------- Row 1 --------
\includegraphics[width=0.32\textwidth]{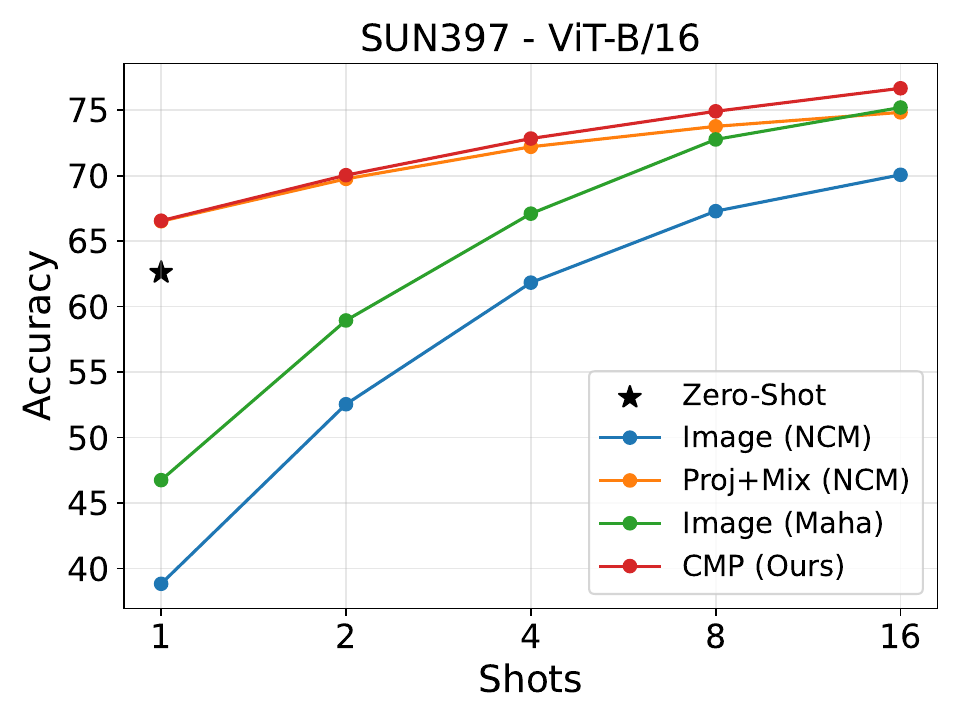}\hfill
\includegraphics[width=0.32\textwidth]{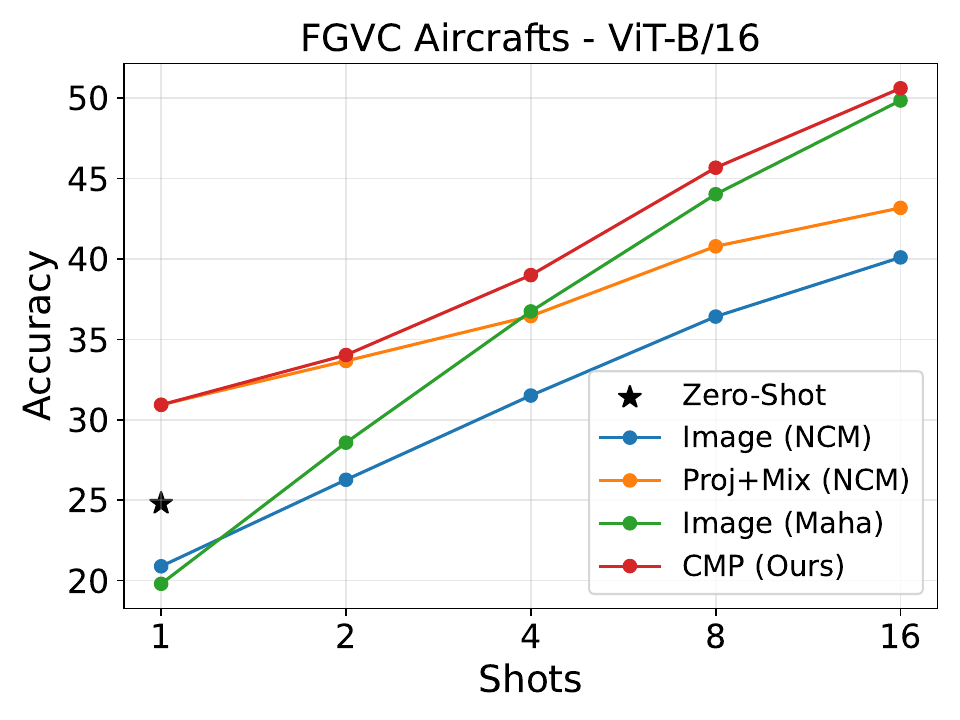}\hfill
\includegraphics[width=0.32\textwidth]{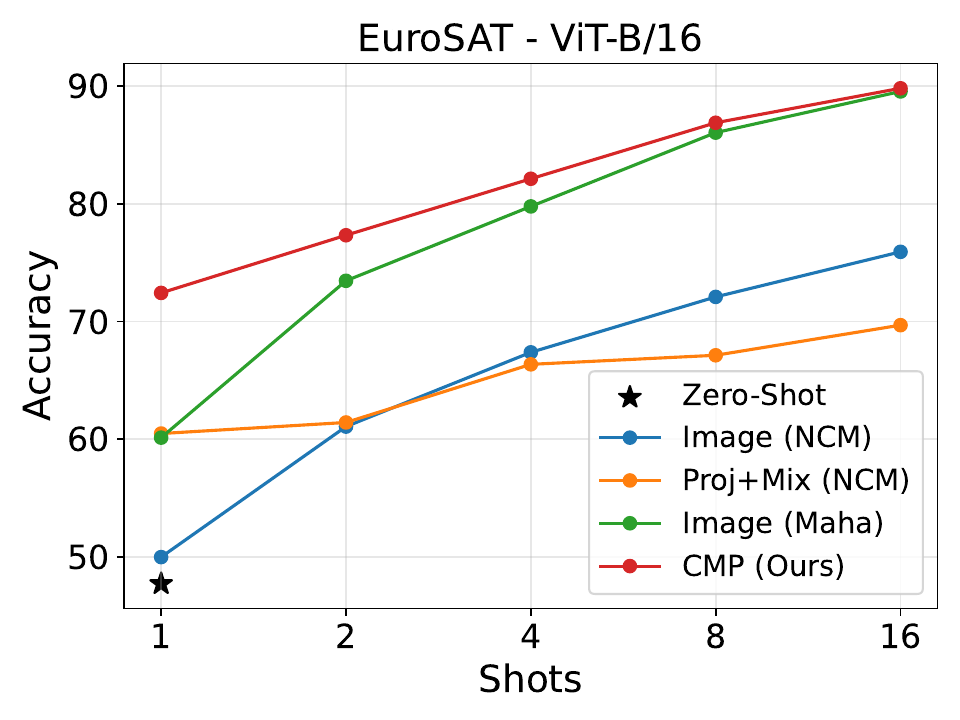}

\vspace{-5pt}
\caption{Impact of different classifiers across 3 datasets and varying shots. The proposed method CMP combines Proj+Mix (NCM) and Image (Maha).
}
\label{fig:classifiers}
\vspace{-15pt}
\end{figure}

% In the document
\begin{figure}[t]
\centering

% -------- Row 1 --------
\includegraphics[width=0.32\textwidth]{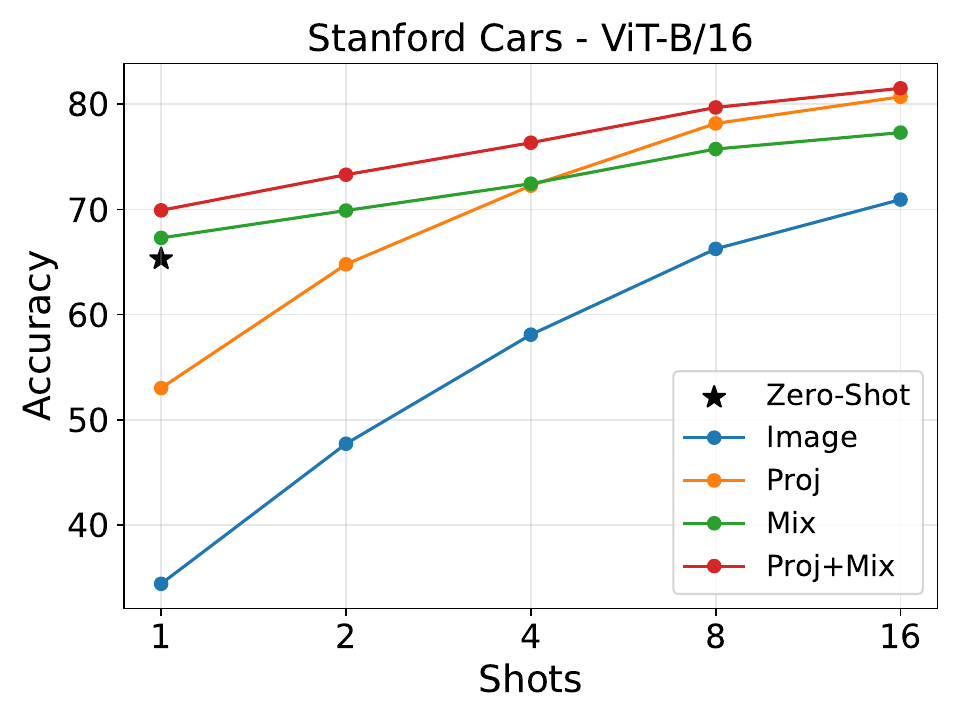}\hfill
\includegraphics[width=0.32\textwidth]{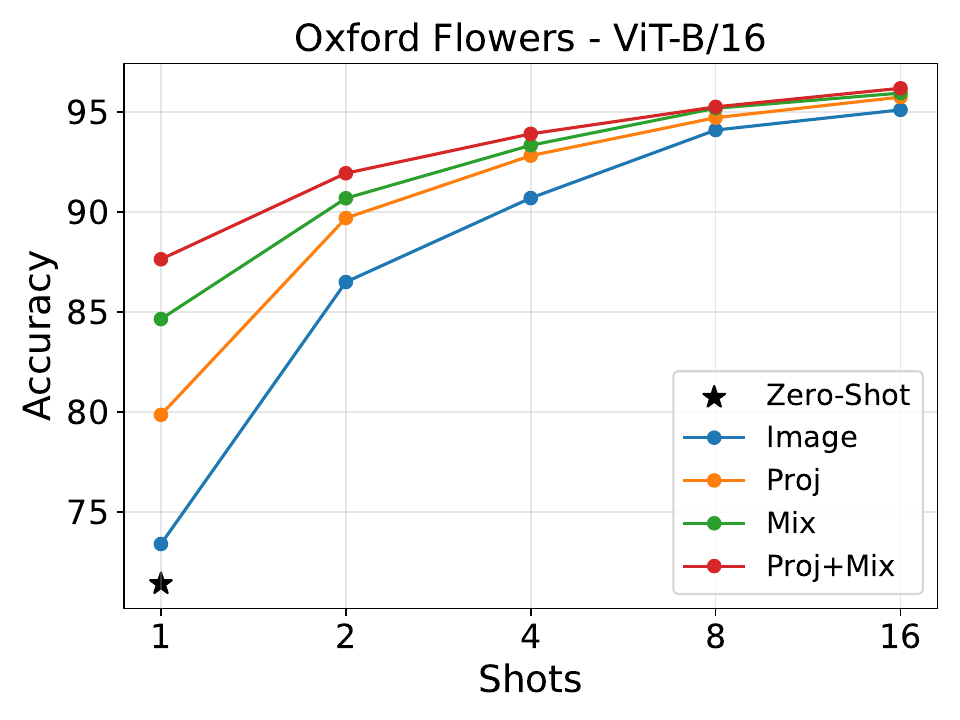}\hfill
\includegraphics[width=0.32\textwidth]{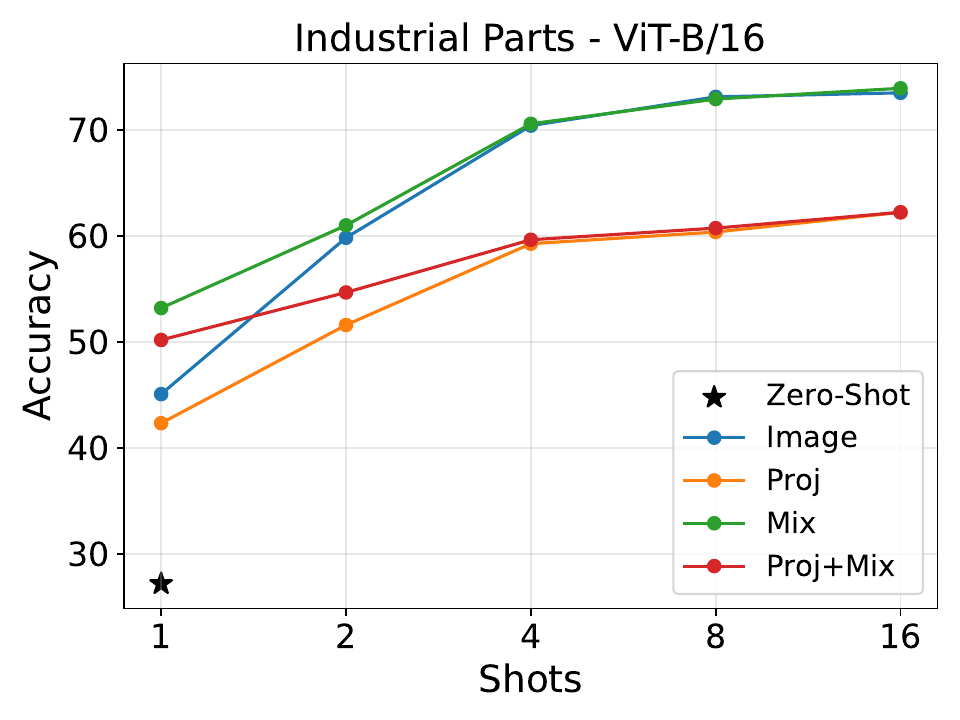}

\vspace{-5pt}
\caption{Ablations of using NCM classifiers with image prototypes (Image), image prototypes in task-semantic subspace (Proj), mixed prototypes in the full image space (Mix) and mixed prototypes in the task-semantic subspace (Proj+Mix).
}
\label{fig:classifiers2}
\vspace{-5pt}
\end{figure}

\subsection{Analysis and Ablation of Classifiers}
We analyze the impact of different classifiers for 3 datasets using CLIP-ViT-B/16 in~\cref{fig:classifiers}. 
On the standard datasets, we observe a higher alignment (mean of principal angle cosines $\mu(S_M) > 0.4$) while on the OOD datasets, $\mu(S_M) < 0.4$. This alignment affects the performance of the Proj+Mix classifier. We present $\mu(S_M)$ scores for all datasets in~\cref{app:datasets}.
On SUN397 having $\mu(S_M) \in [0.84, 0.85]$ across shots and FGVC Aircrafts $\big(\mu(S_M) \in [0.43, 0.45]\big)$, Proj+Mix significantly improves accuracy over Zero-Shot CLIP and NCM on image prototypes, thus validating the impact of using only the task-semantic subspace. In EuroSAT with poor alignment $\big(\mu(S_M) \in [0.2, 0.23]\big)$, Proj+Mix performs rather poorly and thus depends more on the Mahalanobis image classifier. 
Across all datasets, the anisotropic Mahalanobis classifier (Maha) is better than the isotropic NCM classifier. As the number of shots increases to 16, we see an improvement in the Image (Maha) classifier performance due to better covariance estimates leading to overall better performance of CMP which combines Proj+Mix (NCM) and Image (Maha).

In~\cref{fig:classifiers2}, we present the ablations of using NCM classifiers with Image, Proj, Mix and Proj+Mix variants of prototypes. For datasets with good alignment like Stanford Cars $\big(\mu(S_M) \in [0.59, 0.61]\big)$ and Oxford Flowers $\big(\mu(S_M) \in [0.5, 0.52]\big)$, we observe similar trends - Proj consistently improves over naive Image classifiers, Mix improves over Proj and Image with decreasing impact as the number of shots increases, and finally Proj+Mix improves Mix, Proj and Image classifiers. For OOD datasets like Industrial Parts $\big(\mu(S_M) \in [0.19, 0.23]\big)$, we observe that Proj+Mix improves over Proj while Image and Mix performs better than Proj+Mix. This indicates that the full image space is more useful here compared to the task-semantic subspace due to the poor alignment.

In~\cref{app:ablations} we provide more analysis of classifiers across all datasets, the impact of the number of classes and analysis of hyperparameter selection.

\subsection{Comparison with Training-based Methods} We compare with MaPLe~\citep{khattak2023maple}, which learns multi-modal prompts, and CLIP-LoRA~\citep{zanella2024cliplora}, which adds LoRA modules for few-shot adaptation. We show in~\cref{tab:all2all} that the proposed training-free method CMP outperforms training-based MaPLe by 12\% on 4-shot and 9.4\% on 16-shot settings. Training-free CMP also outperforms CLIP-LoRA on 4-shot setting while performing competitively on 16-shots. 
We also demonstrate that our method can be easily added as a plug-in on top of the embedding space learned using these methods. Using our method with MaPLe improves performance by 11\% on 4-Shot and 9.4\% in 16-Shot settings on average. Similarly, we see an improved performance of 3.3\% on 4-Shot and 1.7\% on 16-Shot on average using CMP on the adapted model from CLIP-LoRA.

\begin{table*}[t!]
    \def\arraystretch{1.1}
    \centering
    \footnotesize
    \caption{Performance comparison of proposed training-free (TF) method CMP with training-based methods using CLIP-ViT-B/16 backbone. Best results denoted in \textbf{bold} and second best is \underline{underlined}.}
    \vspace{-5pt}
    \begin{adjustbox}{max width=\textwidth}
    \begin{tabular}{clccccccccccccccc|c}
    \toprule
     & & \multicolumn{10}{c}{\cellcolor{blue!10}Standard Datasets} & \multicolumn{5}{c}{\cellcolor{red!15}OOD Datasets} & \\
\cmidrule(lr){3-12} \cmidrule(lr){13-17}
     Shot & Method & SUN & CARS & FOOD & CAL & FLWR & ImgNet & UCF & INSCT & AIR & PETS & YOGA & DTD & LEGO & ESAT  & INDS  & Avg \\
    \midrule

& \emph{Zero-Shot} &  62.6 & 65.3 & 86.1 & 92.9 & 71.4 & 68.8 & 66.8 & 35.4 & 24.8 & 89.1 & 6.6 & 44.4 & 8.9 & 47.7 & 27.2 & 53.2  \\
\cmidrule(lr){1-18}

\cellcolor{white}{}  & CMP (TF) & 72.8 & \underline{77.2} & \textbf{86.5} & 95.2 & \underline{95.3} & 70.5 & 78.6 & \underline{76.8} & 39.0 & 92.2 & \underline{25.2} & 64.3 & \textbf{65.3} & 82.1 & 87.1 & \underline{73.9} \\
& MaPLe  & 71.6 & 69.8 & \textbf{86.5} & 94.4 & 84.1 & 70.5 & 76.4 & 41.5 & 29.6 & \textbf{93.2} & 7.4 & 58.4 & 21.6 & 65.3 & 57.6 & 61.9 \\
% \hlrow  
\graya \cellcolor{white}{4}  &  + CMP & \underline{74.4} & 76.6 & \underline{86.3} & \underline{95.3} & 94.9 & \underline{71.7} & 80.4 & 76.7 & \underline{39.8} & \underline{92.7} & 20.8 & \underline{65.0} & 51.4 & 82.2 & 85.2 & 72.9 \\ 

& CLIP-LoRA & 73.4 & 76.7 & 83.5 & 95.2 & 93.1 & 71.6 & \underline{81.9} & 71.9 & 37.3 & 91.9 & 23.4 & 63.6 & 53.6 & \underline{84.9} & \underline{87.8} & 72.7 \\

\graya 
\cellcolor{white}{}  &  + CMP & \textbf{75.9} & \textbf{80.5} & 84.4 & \textbf{95.9} & \textbf{96.3} & \textbf{72.7} & \textbf{84.0} & \textbf{83.4} & \textbf{42.4} & 92.6 & \textbf{31.1} & \textbf{69.4} & \underline{55.2} & \textbf{88.1} & \textbf{88.4} & \textbf{76.0} \\

 \\ [-6ex] \\ 
    \cmidrule(lr){1-18}
    
& CMP (TF) & 76.7 & 83.6 & 87.4 & 96.2 & 98.1 & 72.5 & 83.8 & 87.7 & 50.6 & \underline{94.0} & 41.8 & 72.7 & \textbf{85.4} & 89.8 & 93.5 & 80.9 \\

& MaPLe  & 74.4 & 74.7 & \underline{87.5} & 95.3 & 93.7 & 71.9 & 80.9 & 67.1 & 36.0 & 93.5 & 15.2 & 67.6 & 42.4 & 85.1 & 86.6 & 71.5 \\
% \hlrow  
\graya 
\cellcolor{white}{16} &  + CMP & 75.5 & \underline{85.6} & \textbf{87.7} & 96.2 & \underline{98.4} & 72.2 & 84.7 & 88.9 & 51.7 & \textbf{94.1} & 43.3 & \underline{74.1} & 76.0 & 91.9 & \underline{93.7} & 80.9 \\

& CLIP-LoRA  & \underline{76.9} & 85.1 & 85.3 & \underline{96.5} & 97.9 & \underline{73.5} & \underline{86.6} & \underline{90.9} & \underline{54.9} & 93.0 & \underline{54.6} & 72.5 & 80.4 & \underline{92.3} & \textbf{96.1} & \underline{82.4} \\

% \hlrow 
\graya 
\cellcolor{white}{} &  + CMP & \textbf{78.4} & \textbf{87.7} & 86.6 & \textbf{96.8} & \textbf{98.6} & \textbf{74.7} & \textbf{88.2} & \textbf{92.0} & \textbf{59.9} & 93.7 & \textbf{59.3} & \textbf{75.5} & \underline{81.3} & \textbf{93.1} & \textbf{96.1 } & \textbf{84.1} \\

\bottomrule
    \end{tabular}
    \end{adjustbox}
\label{tab:all2all}
\vspace{-5pt}
\end{table*}

\section{Conclusions}
In this work, we introduced a training-free approach to few-shot classification with vision-language models. Our bias–variance analysis shows that mixing image and text prototypes improves classification, but remains suboptimal because the two modalities are only partially aligned. To isolate task-specific semantics, our semantic subspace classifier projects image prototypes onto the principal directions of the text embedding space and mixes only their task-semantic components with text prototypes. Because cross-modal alignment varies across datasets and the image space contains additional discriminative variance, we complement this classifier with an image-specific classifier that models class covariances. Fusing the two classifiers leverages the complementary strengths of both modalities. Across 15 benchmarks, our approach consistently outperforms existing training-free baselines and integrates seamlessly with models updated through training-based methods.

\minisection{Limitations and Future Work.}
Our approach relies on the cross-modal alignment of the pre-trained CLIP model. For out-of-distribution datasets with 
severe domain shift or inadequate text representations (e.g. EuroSAT, Lego Bricks), the task-semantic projection provides limited benefit.
Additionally, our image-specific classifier is less effective in low-shot scenarios because robust covariance estimation is unstable. 
Moving forward, we plan to explore the integration of parameter-efficient fine-tuning to explicitly encourage and improve subspace alignment for challenging downstream tasks.

\bibliography{iclr2027_conference}
\bibliographystyle{iclr2027_conference}

%%%%%%%%%%%%%%%%%%%%%%%%%%%%%%%%%%%%%%%%%%%%%%%%%%%%%%%%
\clearpage
\begin{appendices}
\crefalias{section}{appendix}
\section{Mean Squared Error Derivations}
\label{app:mse-derivations}

We first review the Mean Squared Error (MSE) for a generic estimator, then extend the bias-variance decomposition of MSE to $\mathbb{R}^d$ space. Finally, we apply this analysis to the mixed prototype estimator and to the mixed prototype estimator confined to the semantic subspace.

\subsection{Mean Square Error for a generic estimator}

Recall that for a generic estimator $\hat{\theta}:\Omega \rightarrow  \mathbb{R}$, the MSE with respect to the true value $\theta^*$ is defined as
\begin{equation}
    \mathrm{MSE}(\hat{\theta}, \theta^*) 
    =  \E_{X\sim P_{\theta^*}} \!\left[(\hat{\theta}(X) - \theta^*)^2\right]
    = \mathrm{Bias}^2(\hat{\theta}, \theta^*) 
    + \mathrm{Var}(\hat{\theta}),
\end{equation}
where
\begin{equation}
\label{eq:bias}
    \mathrm{Bias}(\hat{\theta}, \theta^*) 
    =  \E_{X\sim P_{\theta^*}} [\hat{\theta}(X)] - \theta^*,
\end{equation}
and
\begin{equation}
\label{eq:variance}
    \mathrm{Var}(\hat{\theta}) 
    =  \E_{X\sim P_{\theta^*}} \!\left[
        (\hat{\theta}(X) -  \E_{X \sim P_{\theta^*}}[\hat{\theta}(X)])^2
    \right].
\end{equation}
Here $X \sim P_{\theta^*}$ denotes samples drawn from the data distribution
under the true parameter $\theta^*$.
From now on, all expectations are taken with respect to
$X \sim P_{\theta^*}$, and we omit this dependence for notational simplicity.

\subsection{Bias--Variance Decomposition in $\mathbb{R}^d$}

We now extend the bias–variance decomposition to the case of
vector-valued estimators $\hat{\theta} \in \mathbb{R}^d$.
The bias is defined as
\begin{equation}
\label{eq:vector-bias}
    \mathrm{Bias}(\hat{\theta}, \theta^*)
    =
     \E[\hat{\theta}] - \theta^*,
\end{equation}
which is now a vector in $\mathbb{R}^d$. The squared bias is :

\begin{equation}
     \mathrm{Bias}^2(\hat{\theta}, \theta^*) =   ( \E[\hat{\theta}] - \theta^*)^{\top}(  \E[\hat{\theta}] - \theta^*)= \| \E[\hat{\theta}] - \theta^* \|^2.
\end{equation}
The covariance matrix of $\hat{\theta}$ is
\begin{equation}
    \Cov(\hat{\theta})
    =
     \E
    \left[
        (\hat{\theta}- \E[\hat{\theta}])
        (\hat{\theta}- \E[\hat{\theta}])^\top
    \right].
\end{equation}
The corresponding variance is defined as:
\begin{equation}
    \mathrm{Var}(\hat{\theta}) = \operatorname{tr}(\Cov(\hat{\theta})).
\end{equation}
Consequently, the Mean Squared Error with the squared Euclidean norm is defined as:
\begin{equation}
    \mathrm{MSE}(\hat{\theta}, \theta^*)
    =
     \E\!\left[
        \|\hat{\theta} - \theta^*\|^2
    \right] = \mathrm{Bias}^2(\hat{\theta},\theta^*)
  + \mathrm{Var}(\hat{\theta}) =   \| \E[\hat{\theta}] - \theta^*\|^2
    +
    \operatorname{tr}\!\left(
        \Cov(\hat{\theta})
    \right).
\end{equation}

\subsection{Bias--Variance Decomposition for Mixed Prototype Estimator}

Defining $\mu_i^{*}$ as the true population image class mean for a generic class $c$ (we omit the class index $c$ to avoid cluttered notation) and $\hat{\mu}_i$ as the empirical sample mean computed from $n$ samples, we get:
\begin{equation}
\label{eq:mean_expectation}
    \E[\hat{\mu}_i] = \mu_i^{*}, \quad \Cov[\hat{\mu}_i] = \frac{\Sigma_i^*}{n}
\end{equation}
where $\Sigma_i^*$ is the population covariance for class $c$. We assume that the text prototype $\mu_t^{*}$ for class $c$ is deterministic, since it comes from a fixed prompt template fed into the text encoder. Hence:
\begin{equation}
\label{eq:determ_text}
    \E[\mu_t^{*}] = \mu_t^*, \quad \Cov [\mu_t^{*}] = 0.
\end{equation}
The mixed prototype estimator is defined as:
\begin{equation}
    \hat{\mu}_{\text{mix}} = \lambda \hat{\mu}_i + (1-\lambda) \mu_t^*.
\end{equation}
We now analyze the bias–variance decomposition of the mixed estimator.

\minisection{Bias of the mixed estimator.} By computing the expectation we have:
\begin{equation}
   \E[\hat{\mu}_{\text{mix}}] 
   = \E[\lambda \hat{\mu}_i + (1-\lambda) \mu^*_t] 
   = \lambda \E[\hat{\mu}_i] + (1-\lambda)\E[\mu^*_t] 
   = \lambda \mu_i^{*} + (1-\lambda)\mu_t^{*},
\end{equation}
where in the last step we have used that $\E[\hat{\mu_i}] = \mu_i^*$ and $\E[\mu_t^*] = \mu_t^*$. The squared bias is given by:
\begin{equation}
\mathrm{Bias}^2(\hat{\mu}_{\text{mix}}, \mu_i^*) = \| \E[\hat{\mu}_{\text{mix}}] - \mu_i^* \|^2 = \left\| \lambda \mu_i^{*} + (1-\lambda)\mu_t^{*} - \mu_i^* \right\|^2 
\end{equation}

\minisection{Covariance of the mixed estimator.} Assuming that the image and text prototypes are independent, 
$\Cov[\hat{\mu}_i, \mu^*_t]=0$ and since we assumed that the text prototype is deterministic ($\Cov[\mu_t^*] = 0$), we obtain:
\begin{equation}
    \Cov[\hat{\mu}_{\text{mix}}] 
    = \lambda^2 \Cov[\hat{\mu}_i] 
    + (1-\lambda)^2 \Cov[ \mu^*_t] = \lambda^2 \Cov[\hat{\mu}_i]  = \lambda^2 \frac{\Sigma_i^*}{n},
\end{equation}
where the last equality follows from~\eqref{eq:mean_expectation}.

\minisection{MSE of the mixed estimator.} The MSE, obtained by summing the squared bias and the variance, is given by:
\begin{equation}
\label{eq:mse-mix-SUPP}
\begin{aligned}
    \text{MSE}(\hat{\mu}_{\text{mix}}, \mu_i^{*})
    =& \left\| \lambda \mu_i^{*} + (1-\lambda)\mu_t^{*} - \mu_i^* \right\|^2 
    + \lambda^2 \frac{1}{n} \text{tr}(\Sigma_i^*) \\
    =& \left\| (\lambda-1) \mu_i^{*} + (1-\lambda)\mu_t^{*}  \right\|^2 + \lambda^2 \frac{1}{n} \text{tr}(\Sigma_i^*) \\=& (1-\lambda)^2 \| \mu_t^* -\mu_i^* \|^2 + \lambda^2 \frac{1}{n} \text{tr} (\Sigma_i^*).
\end{aligned}
\end{equation}

\section{Analysis on Mixing in the full image space}
\label{app:mixing-analysis}

Recall that in~\cref{sec:4}, we defined the \textit{task-semantic projector} onto the $k$-dimensional
principal subspace spanned by the left singular vectors of the text prototype matrix as:
\begin{equation}
    P = U_k U_k^{\top} \in \mathbb{R}^{d \times d}.
\end{equation}
This leads to the task-semantic subspace decomposition of the empirical image prototype $\hat{\mu}_i \in \mathbb{R}^d$:
\begin{equation}
\label{eq:decomposition}
\hat{\mu}_i
=
P \hat{\mu}_i
+
(I - P)\hat{\mu}_i = \mupar  + \muort.
\end{equation}
We decompose the mixed prototype estimator as:
\begin{equation}
\begin{aligned}
    \hat{\mu}_{\text{mix}} 
    =& \lambda\hat{\mu}_i  + (1-\lambda) \mu_t^*
    = \lambda (\mupar  + \muort ) + (1-\lambda) \mu_t^* \\
    =& \mu_t^{*} + \lambda (\mupar - \mu_t^*) + \lambda \muort.
\end{aligned}
\end{equation}

This shows that naive mixing interpolates the task-semantic image component $\mupar$ with the text prototype, while simultaneously scaling the orthogonal image component $\muort$ by $\lambda$. Now we will compute the MSE for the mixed prototype estimator.

Now, by applying the linearity of the expectation and recalling that the text prototype is deterministic (\eqref{eq:determ_text}), we obtain the expectation as follows:
\begin{equation}
   \E[\hat{\mu}_{\text{mix}}] 
   = \E[\mu_t^{*} + \lambda (\mupar -\mu_t^*) + \lambda \muort] 
   = \mu_t^{*} + \lambda (\mu_i^{\shortpara,*} -\mu_t^*) + \lambda \mu_i^{\perp,*}
\end{equation}
where $\mu_i^{\shortpara,*} =\E[\mupar]$ and $\mu_i^{\perp,*}=\E[\muort]$ denote the task-semantic and task-orthogonal components of the population image class mean. The squared bias is given by:
\begin{equation}
\begin{aligned}
    \mathrm{Bias}^2(\hat{\mu}_{\text{mix}}, \mu_i^*) 
    &= || \E[\hat{\mu}_{\text{mix}}] - \mu_i^* ||^2 \\
    &= \left\| \mu_t^{*} + \lambda (\mu_i^{\shortpara,*} -\mu_t^*) + \lambda \mu_i^{\perp,*} - \mu_i^{\shortpara,*} - \mu_i^{\perp,*} \right\|^2 \\
    &= (1-\lambda)^2 \left\| (\mu_t^{*} - \mu_i^{\shortpara,*}) - \mu_i^{\perp,*} \right\|^2. \\
\end{aligned}
\end{equation}
Since $\mu_t^{*}- \mu_i^{\shortpara,*}$ lies in the task-semantic subspace and $\mu_i^{\perp, *}$ lies in its orthogonal complement, the two vectors are orthogonal. Therefore, $\| a - b\|^2= \|a\|^2 + \|b\|^2$ and the bias become:
\begin{equation}
    \mathrm{Bias}^2(\hat{\mu}_{\text{mix}}, \mu_i^*) = (1-\lambda)^2 \left( \|\mu_t^{*} - \mu_i^{\shortpara,*}\|^2 + \|\mu_i^{\perp,*}\|^2 \right).
\end{equation}

Similarly, we obtain the covariance as:
\begin{equation}
    \Cov[\hat{\mu}_{\text{mix}}] 
    = \lambda^2 \Cov[\mupar] 
    + \lambda^2 \Cov[\muort]  = \lambda^2 \frac{\Sigma_i^{\shortpara,*}}{n} + \lambda^2 \frac{\Sigma_i^{\perp,*}}{n}.
\end{equation}

Finally, the MSE decomposition is given by:
\begin{equation}
\mathrm{MSE}(\hat{\mu}_{\text{mix}}, \mu_i^*)
= (1-\lambda)^2
\left(
\|\mu_t^* - \mu_i^{\shortpara, *}\|^2
+
\|\mu_i^{\perp, *}\|^2
\right)
+
\lambda^2 \frac{1}{n}
\left(
\mathrm{tr}(\Sigma_i^{\shortpara,*})
+
\mathrm{tr}(\Sigma_i^{\perp,*})
\right), 
\label{new-mse}
\end{equation}
\noindent 
where $\mu_i^{\shortpara,*}$ and $\mu_i^{\perp,*}$ denote the task-semantic and task-orthogonal components of population image class mean, and $\Sigma_i^{\shortpara, *}$, $\Sigma_i^{\perp, *}$ are the corresponding class covariances in the two subspaces. Notably, the factor $(1-\lambda)^2$ also scales the bias contribution $\|\mu_i^{\perp, *}\|^2$ from the orthogonal component, even though the text prototype contains no information about it. 

Consequently, naive mixing pulls the text-based semantic component toward the text prototype while also shrinking the orthogonal component toward zero. In particular, reducing $\lambda$ to control the high variance of few-shot image prototypes also suppresses image-specific information in the orthogonal subspace, although no text information is available to guide this shrinkage. 
Therefore, mixing with text prototypes in the full image space is sub-optimal and we propose to handle the image-specific information separately while mixing only the task-semantic component of image prototype with the text prototype.

\section{Datasets}
\label{app:datasets}

The datasets used in our experiments include generic object recognition datasets like ImageNet (ImgNet; \citealp{russakovsky2015imagenet}) having 1000 classes and Caltech101 (CAL; \citealp{fei2006one}) having 100 classes, fine-grained image datasets like FGVC Aircrafts (AIR; \citealp{maji13fine-grained}) having 100 classes, Stanford Cars (CARS; \citealp{krause20133d}) having 196 classes, Oxford Pets (PETS; \citealp{parkhi12a}) having 37 classes, Flowers102 (FLWR; \citealp{nilsback2008automated}) having 102 classes, Flying Insects (INSCT; \citealp{kaggle_insect}) having 100 classes and Food101 (FOOD; \citealp{bossard2014food}) having 101 classes, a scene recognition dataset - (SUN397; \citealp{xiao2010sun}) having 397 classes, an action recognition dataset - (UCF101; \citealp{soomro2012ucf101}) having 101 classes. We also include several out-of-distribution datasets like a satellite image dataset - EuroSAT (ESAT; \citealp{helber2019eurosat}) having 10 classes, a texture dataset - Describable Textures (DTD; \citealp{cimpoi14describing}) having 47 classes, a lego bricks dataset (LEGO; \citealp{kaggle_lego}) having 20 classes, a yoga pose dataset (YOGA; \citealp{kaggle_yoga}) having 107 classes and an industrial parts dataset (INDS; \citealp{kaggle_industrial}) having 10 classes. While 11 of these datasets are commonly used in all CLIP-based few-shot classification benchmarks~\citep{zhang2021tip,zhu2023not,wanghard,bendou2025proker}, we consider 4 new datasets (INSCT, YOGA, LEGO and INDS) following a recent work~\citep{wang2026decompose} which uses these datasets to evaluate generalization performance of multimodal LLMs.

\begin{figure}[t]
\centering

\includegraphics[width=0.7\textwidth]{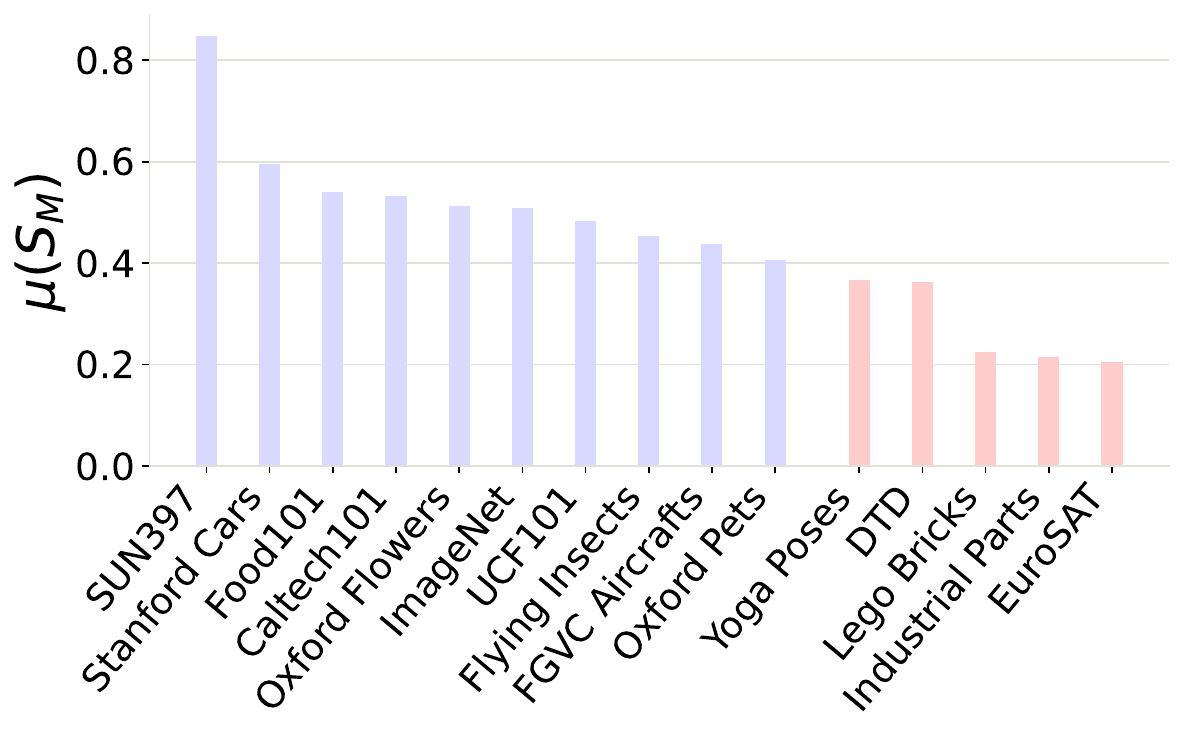}

\vspace{-5pt}
\caption{Analysis of the mean of principal angle cosines $\mu(S_M)$ indicating cross-modal alignment across all datasets in the 4-shot setting using CLIP-ViT-B/16.
}
\label{fig:all_datasets}
% \vspace{-5pt}
\end{figure}

\minisection{Cross-Modal Alignment across Datasets.} We present the mean of the principal angle cosines $\mu(S_M)$ for all datasets in~\cref{fig:all_datasets} using CLIP-ViT-B/16 on the 4-shot setting. These values reflect the existing cross-modal alignment in CLIP for different datasets.

\section{Additional Analysis and Ablations}
\label{app:ablations}

In this section, we provide more analysis and ablations to complement those in the main paper.

\newpage

\begin{wrapfigure}{r}{0.4\textwidth}
  \centering
 \vspace{-10pt}
  \includegraphics[width=\linewidth]{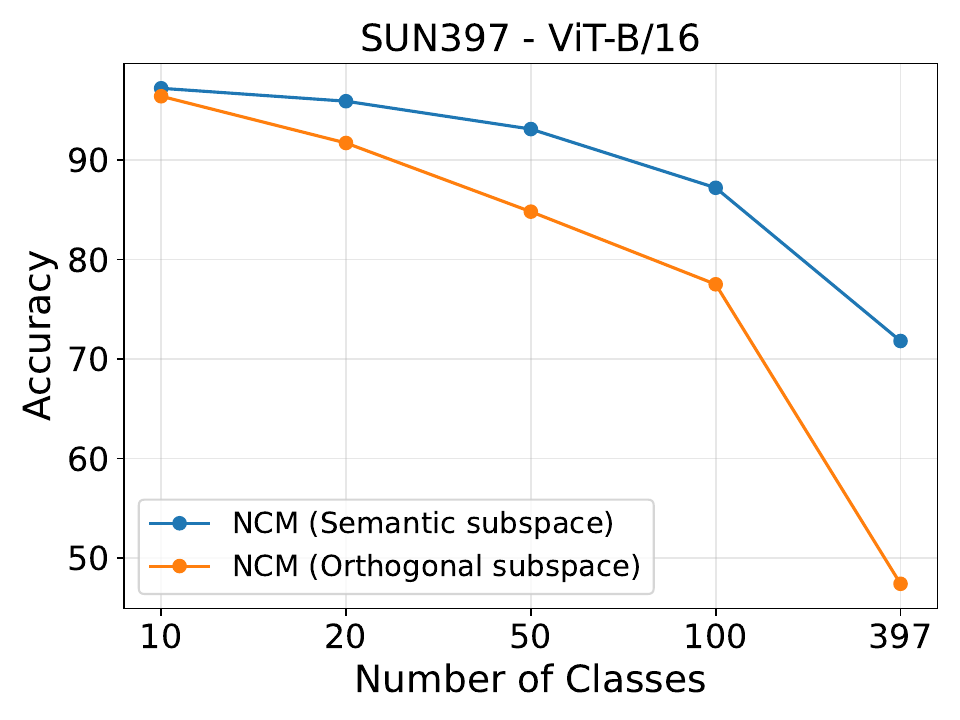}
  \vspace{-10pt}
  \caption{Analysis of the impact of varying subsets of classes on the semantic and orthogonal subspaces.}
  \vspace{-10pt}
  \label{fig:sun}
\end{wrapfigure}

\subsection{Impact of the Number of Classes on the Semantic Projection} 
To study the impact of the number of classes on the proposed semantic subspace decomposition, we consider the SUN397 dataset which has 397 classes and good cross-modal alignment in CLIP. We selected random subsets of classes of varying number of classes in the 16-shot setting and compared the performance of NCM with image prototypes in the task-semantic subspace and in the task-orthogonal subspace as shown in~\cref{fig:sun}. We observe that despite having very few classes and thereby smaller task-semantic component compared to the orthogonal component, NCM in the proposed semantic subspace consistently performs better. This implies that the proposed task-semantic projector identifies the more discriminative subspace irrespective of the number of classes or the rank of the semantic projector.

\subsection{Impact of different classifiers}

\begin{figure}[t]
\centering

% -------- Row 1 --------
\includegraphics[width=0.33\textwidth]{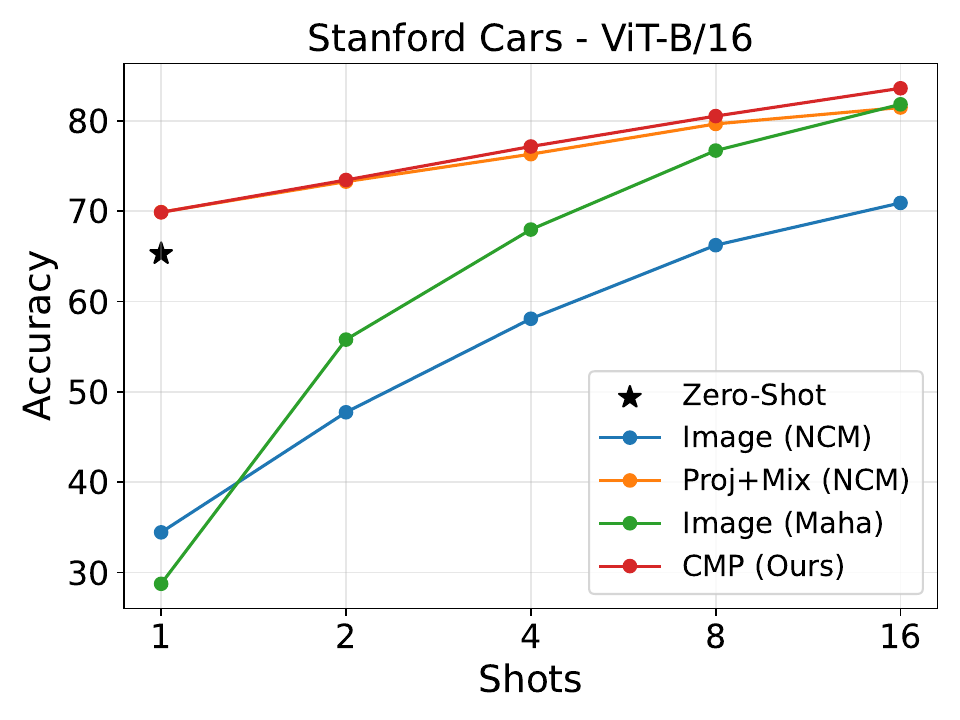}\hfill
\includegraphics[width=0.33\textwidth]{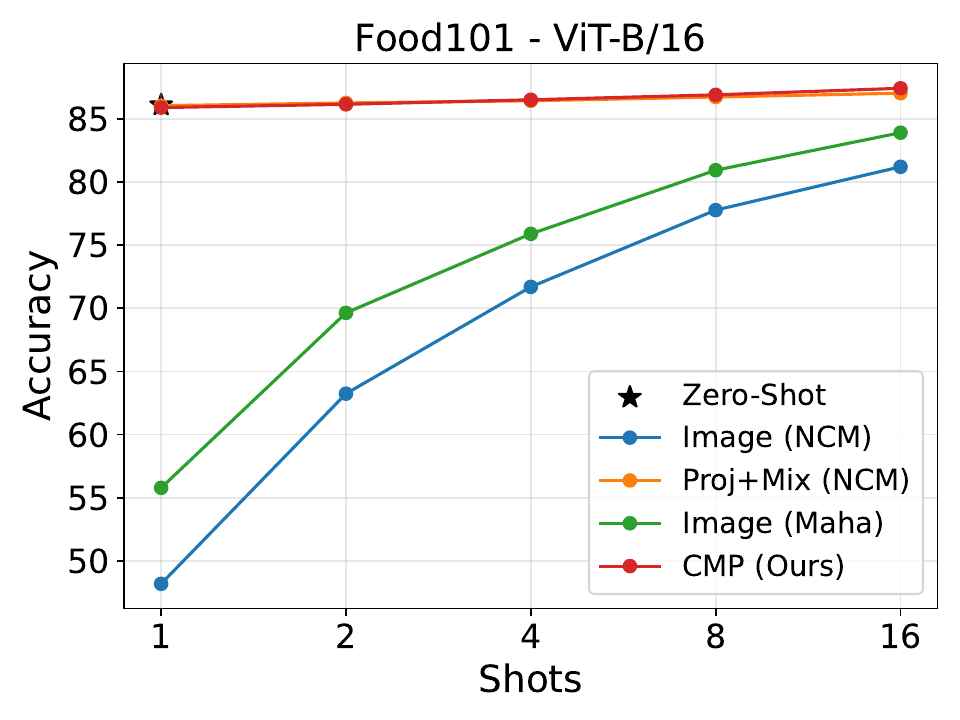}\hfill
\includegraphics[width=0.33\textwidth]{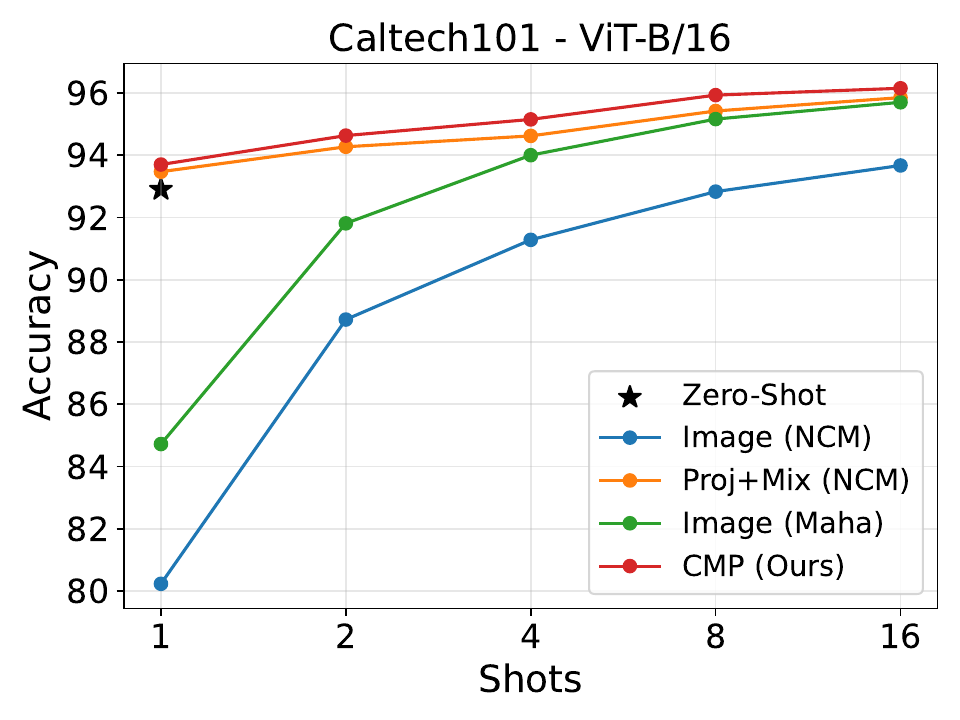}

% -------- Row 2 --------
\includegraphics[width=0.33\textwidth]{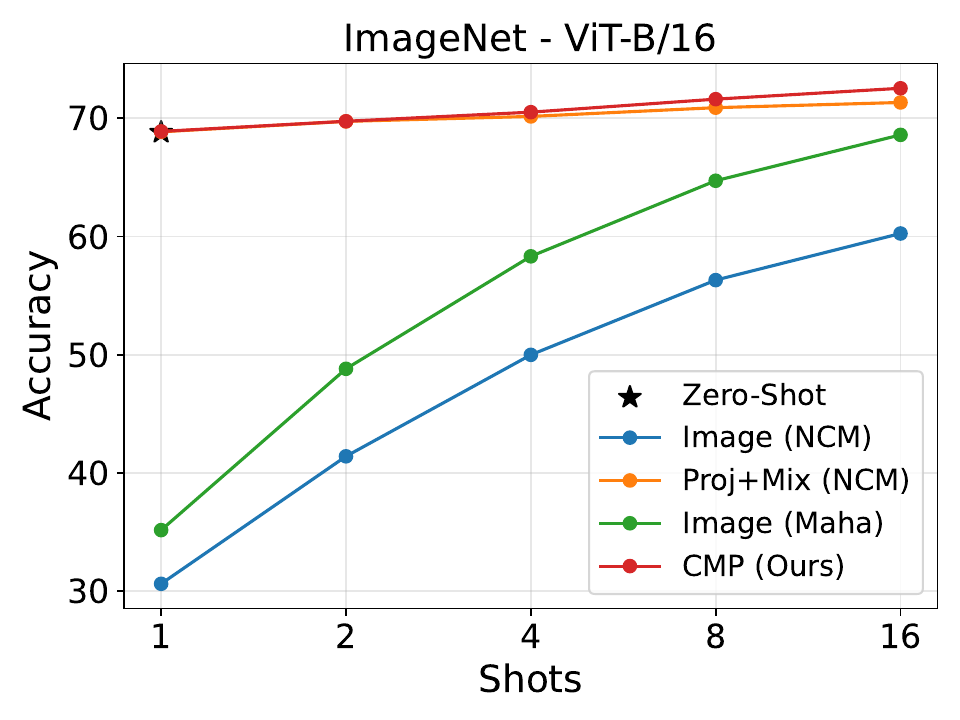}\hfill
\includegraphics[width=0.33\textwidth]{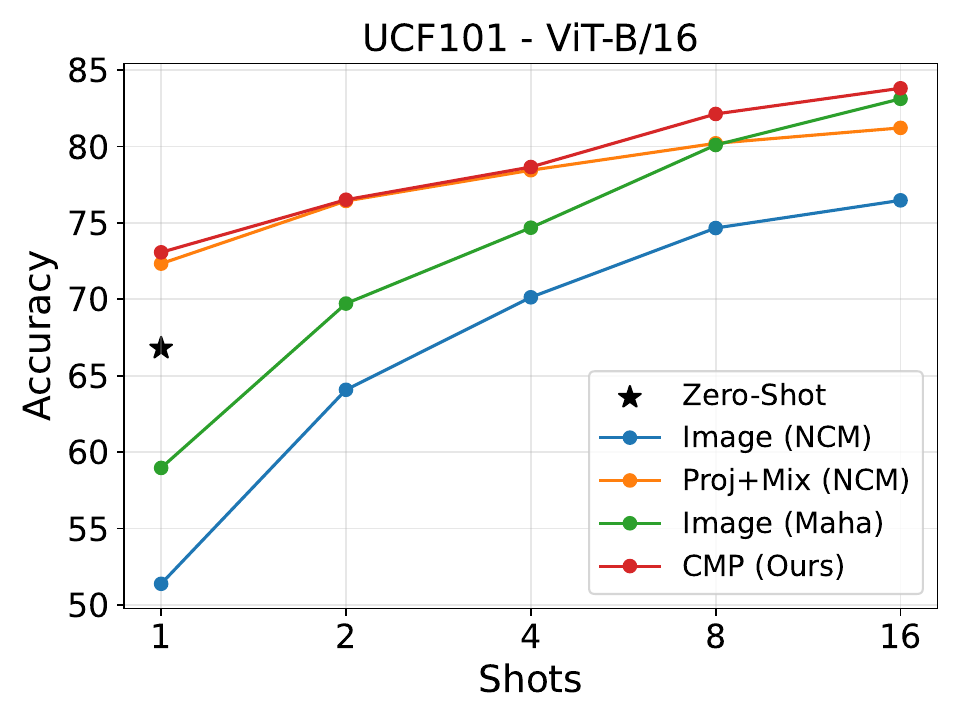}\hfill
\includegraphics[width=0.33\textwidth]{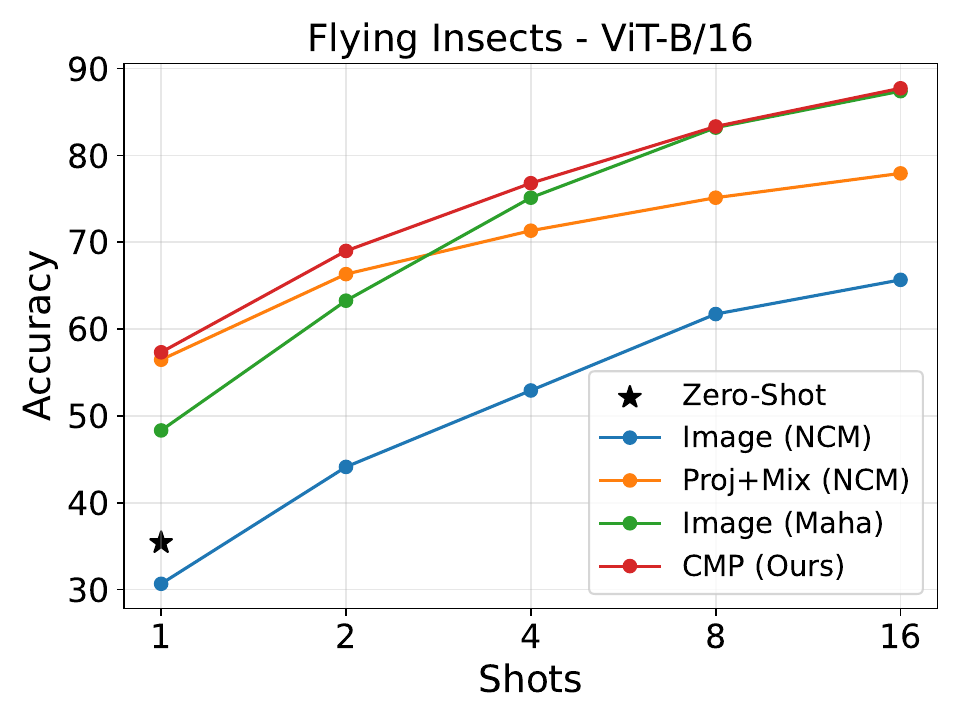}

% -------- Row 3 --------
\includegraphics[width=0.33\textwidth]{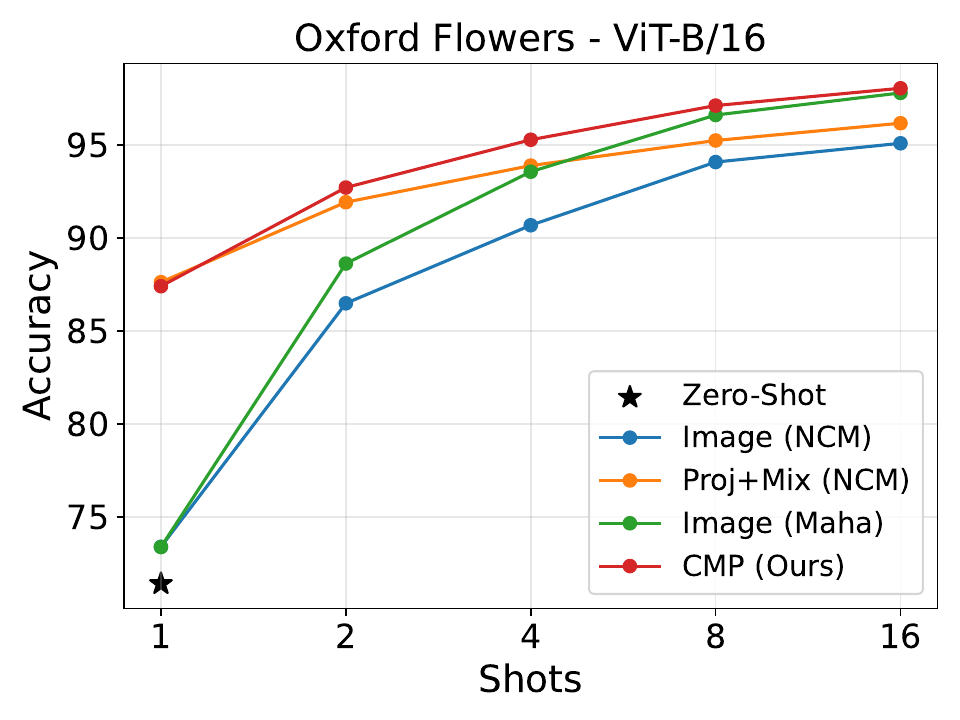}\hfill
\includegraphics[width=0.33\textwidth]{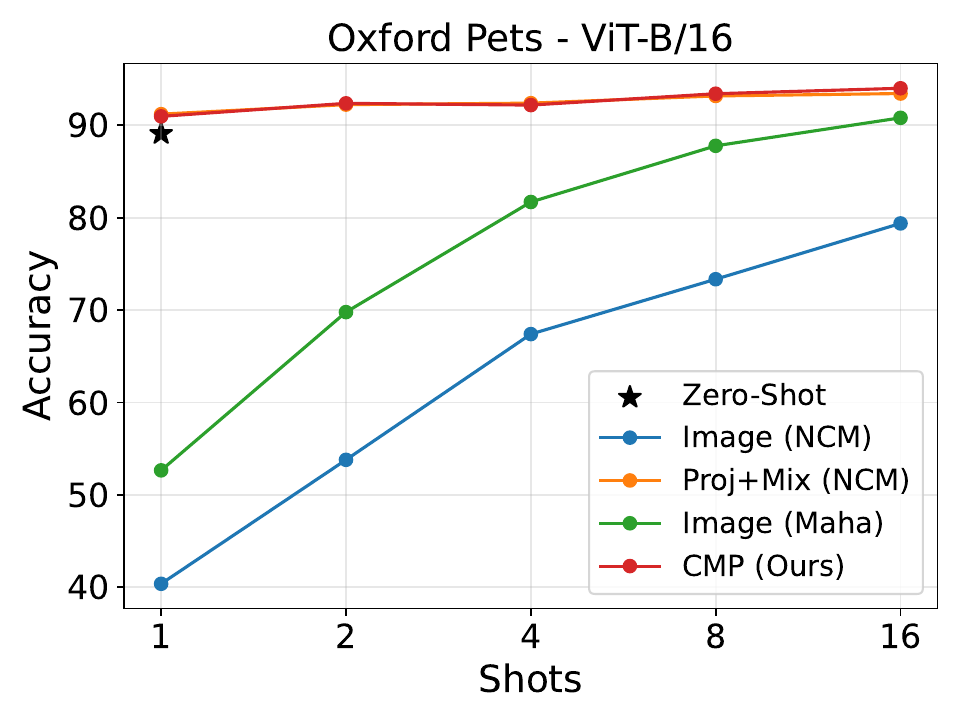}\hfill
\includegraphics[width=0.33\textwidth]{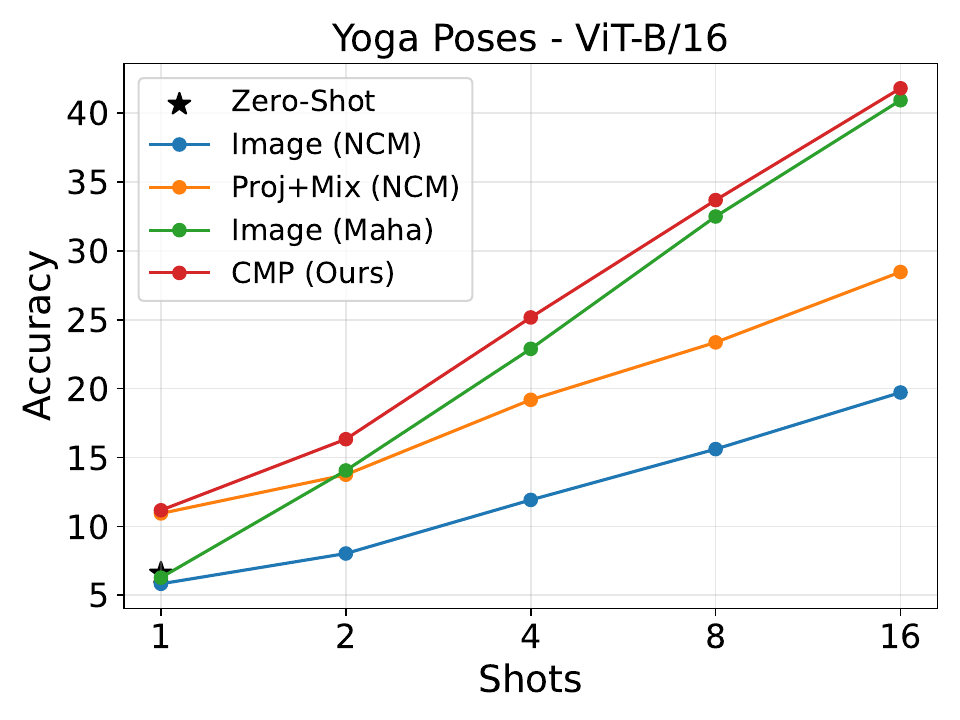}

% -------- Row 4 --------
\includegraphics[width=0.33\textwidth]{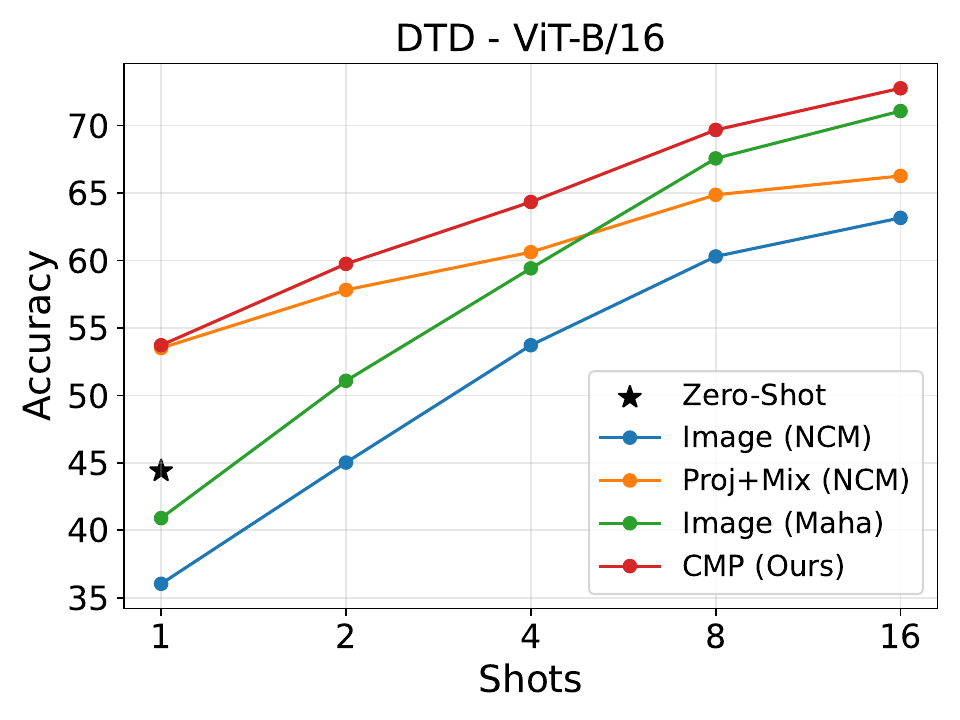}\hfill
\includegraphics[width=0.33\textwidth]{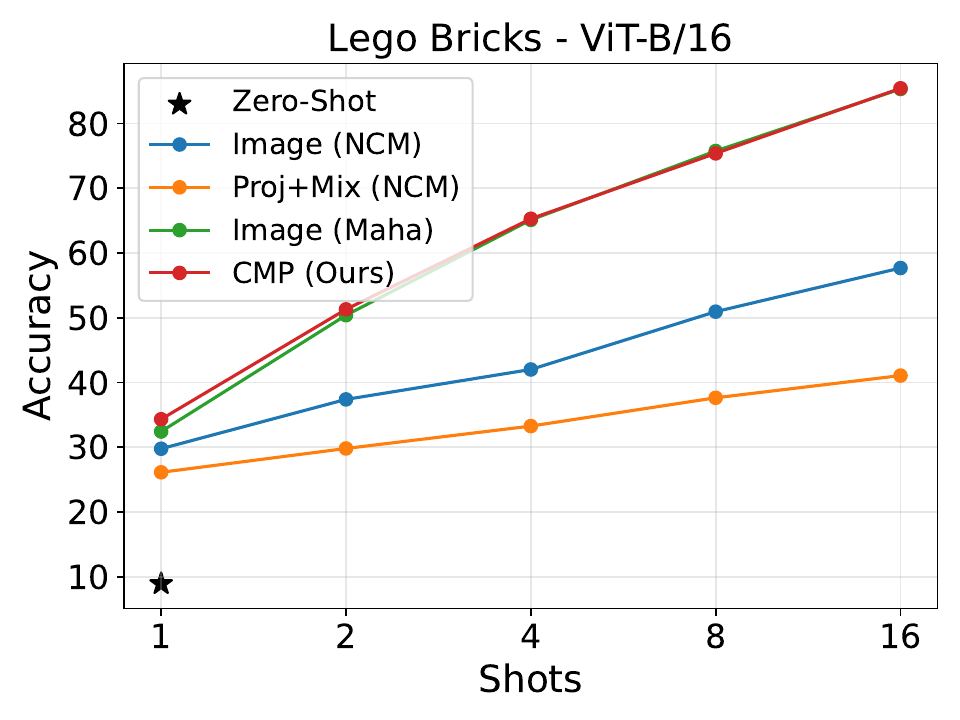}\hfill
\includegraphics[width=0.33\textwidth]{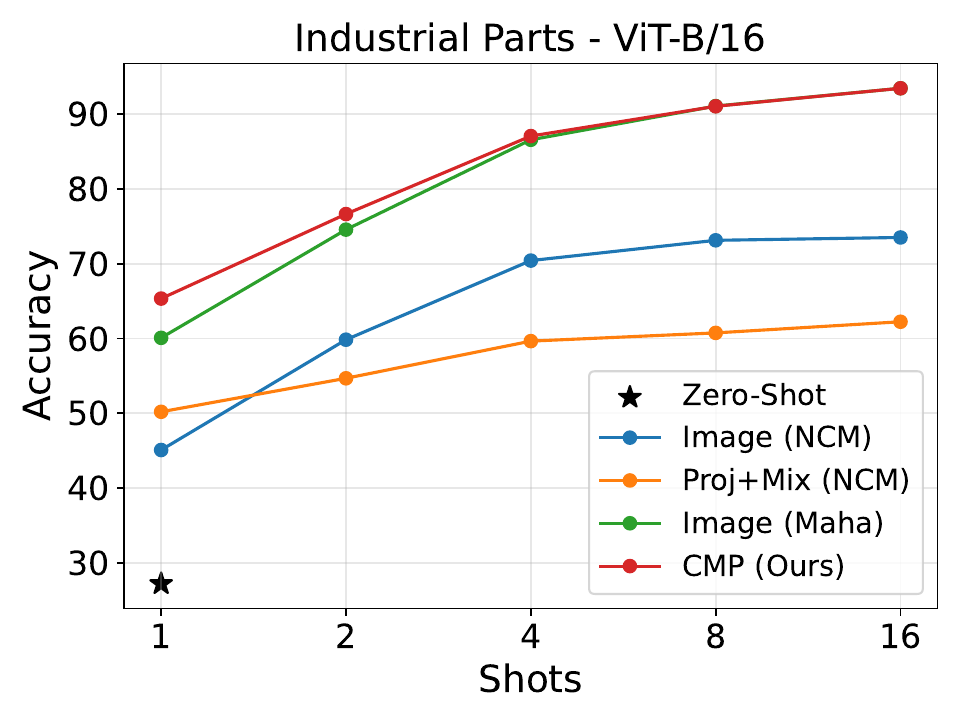}

\caption{Impact of different classifiers across 3 datasets and varying shots on 12 datasets with CLIP-ViT-B/16.. The proposed method CMP combines Proj+Mix (NCM) and Image (Maha).}
\label{fig:classifiers_2}
\vspace{-5pt}
\end{figure}

We demonstrate the impact of different classifiers in~\cref{fig:classifiers_2} for 12 datasets using CLIP-ViT-B/16. We observe some consistent trends across all datasets. Image (Maha) improves significantly over Image (NCM) since it exploits the image covariances. The Proj+Mix (NCM) classifier performs very well in the lower shots for most datasets except the OOD datasets. The proposed method CMP combining Proj+Mix (NCM) and Image (Maha) always outperforms the other classifiers.

\subsection{Impact of Mixing in the Semantic Subspace}

\begin{figure}[t]
\centering

\includegraphics[width=0.33\textwidth]{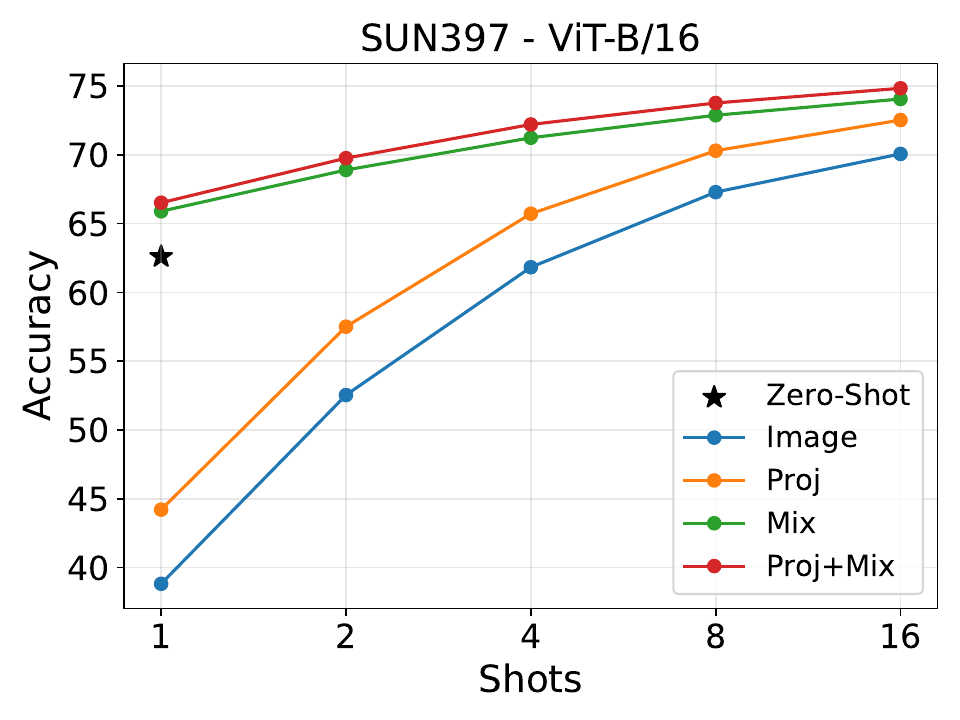}\hfill
\includegraphics[width=0.33\textwidth]{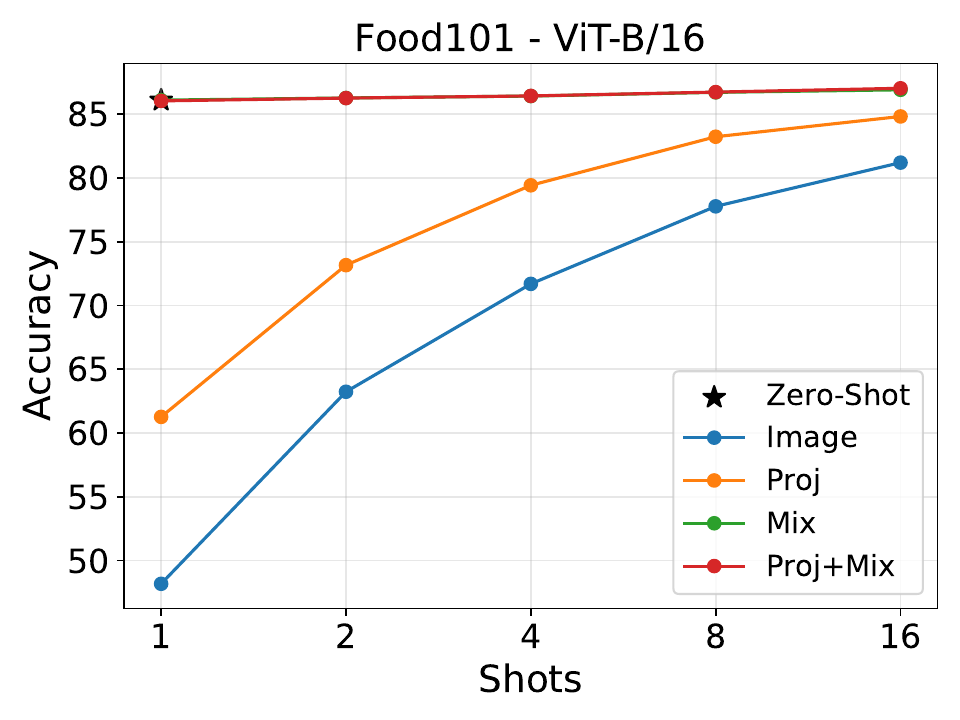}\hfill
\includegraphics[width=0.33\textwidth]{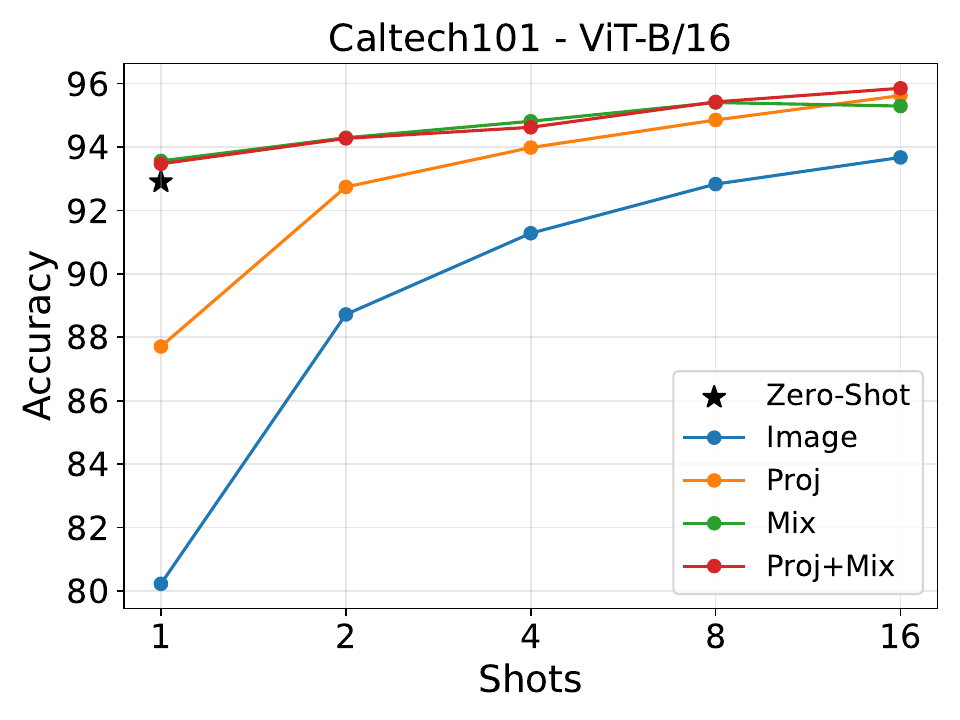}

\includegraphics[width=0.33\textwidth]{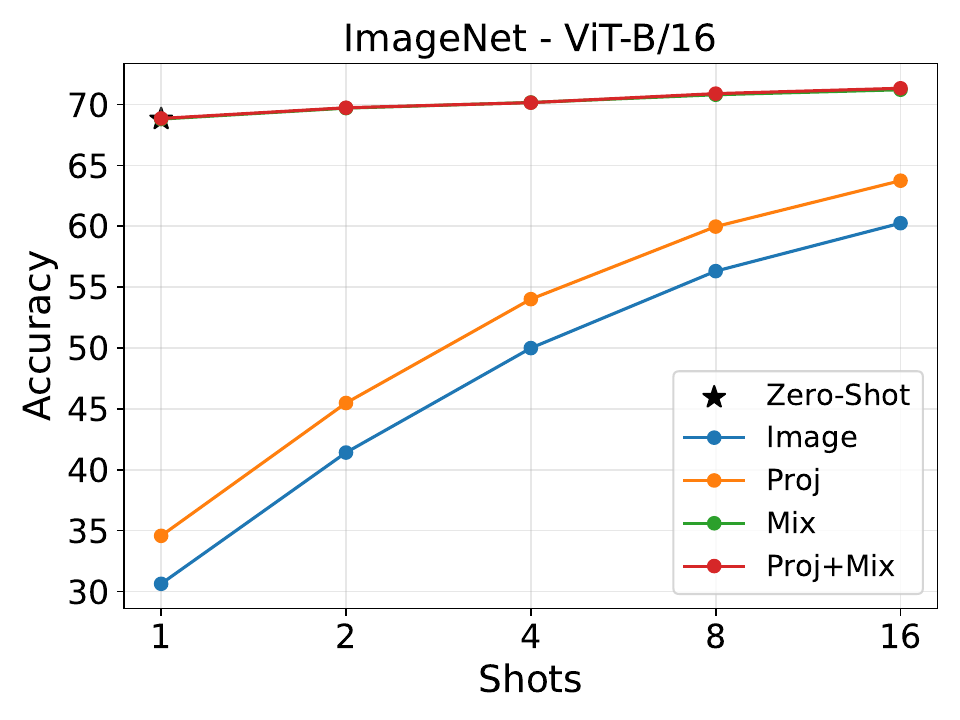}\hfill
\includegraphics[width=0.33\textwidth]{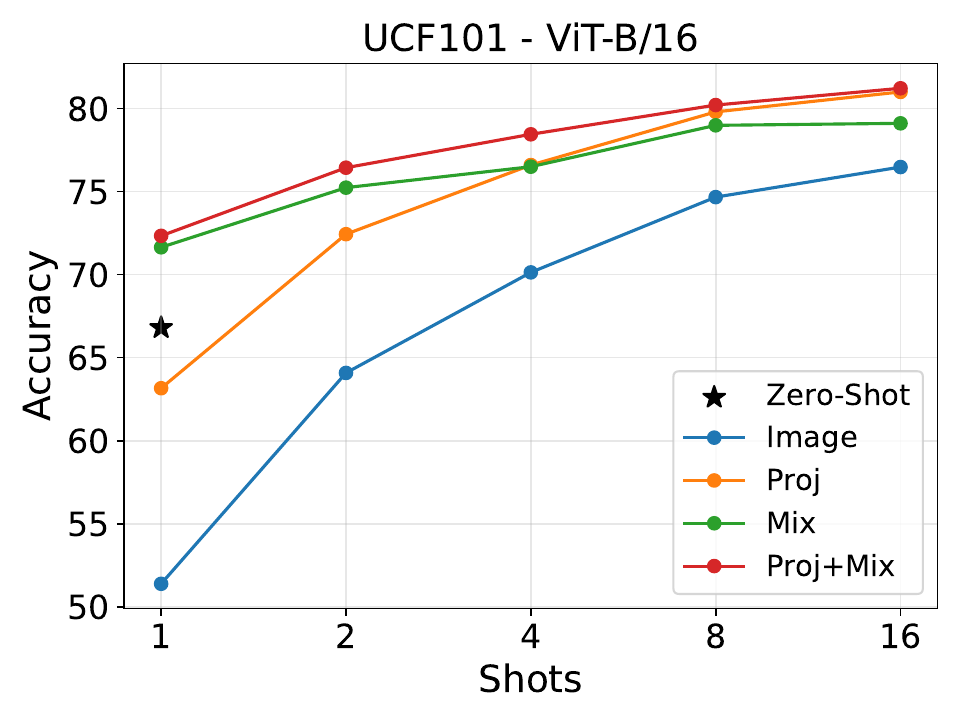}\hfill
\includegraphics[width=0.33\textwidth]{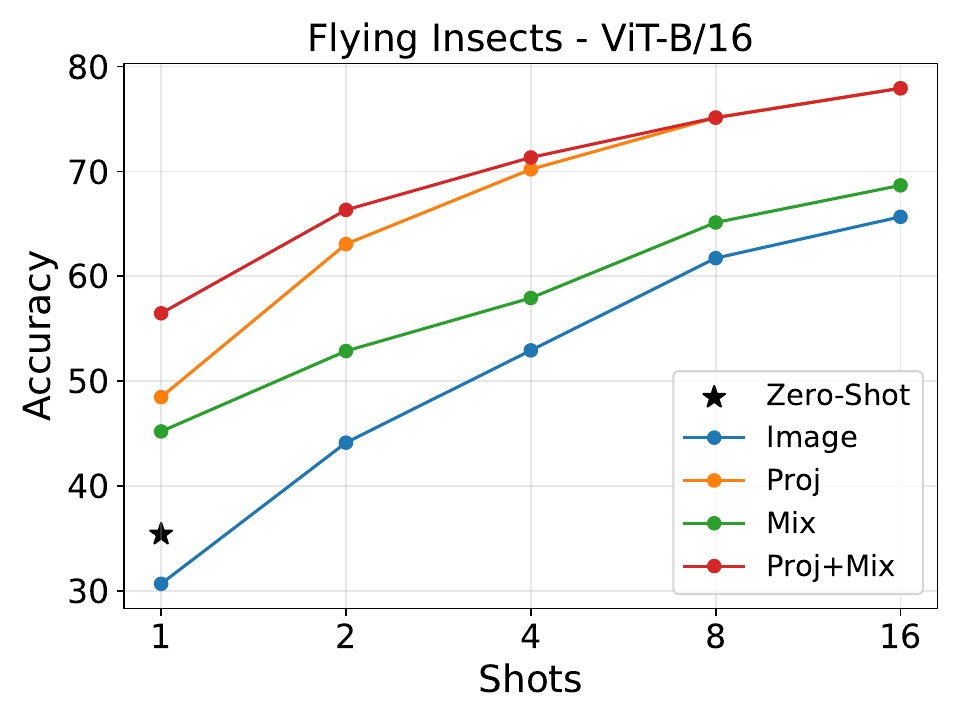}

\includegraphics[width=0.33\textwidth]{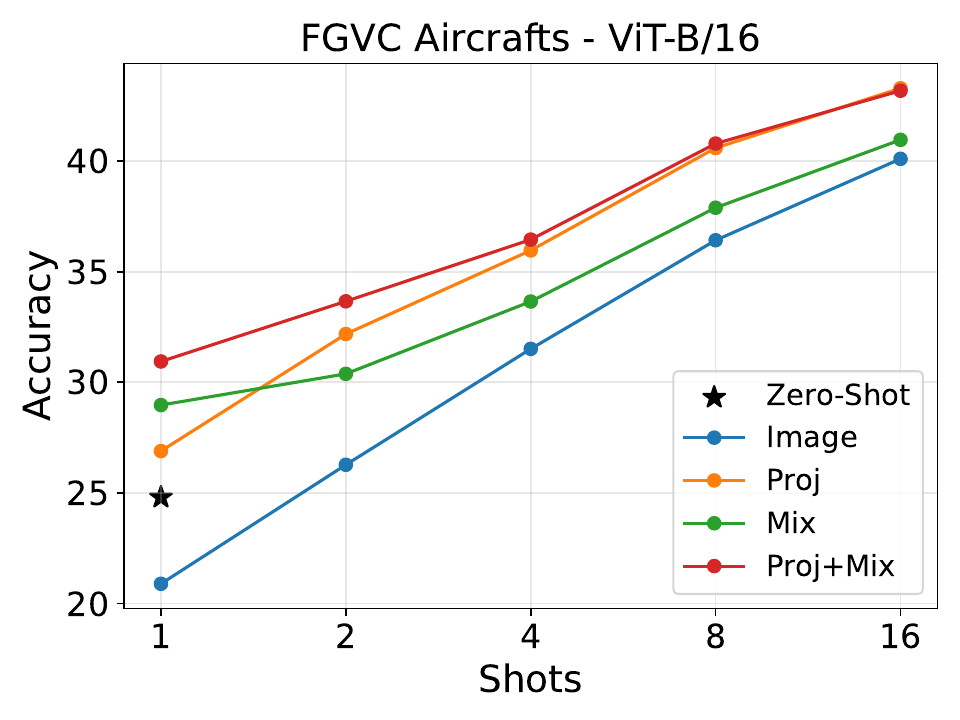}\hfill
\includegraphics[width=0.33\textwidth]{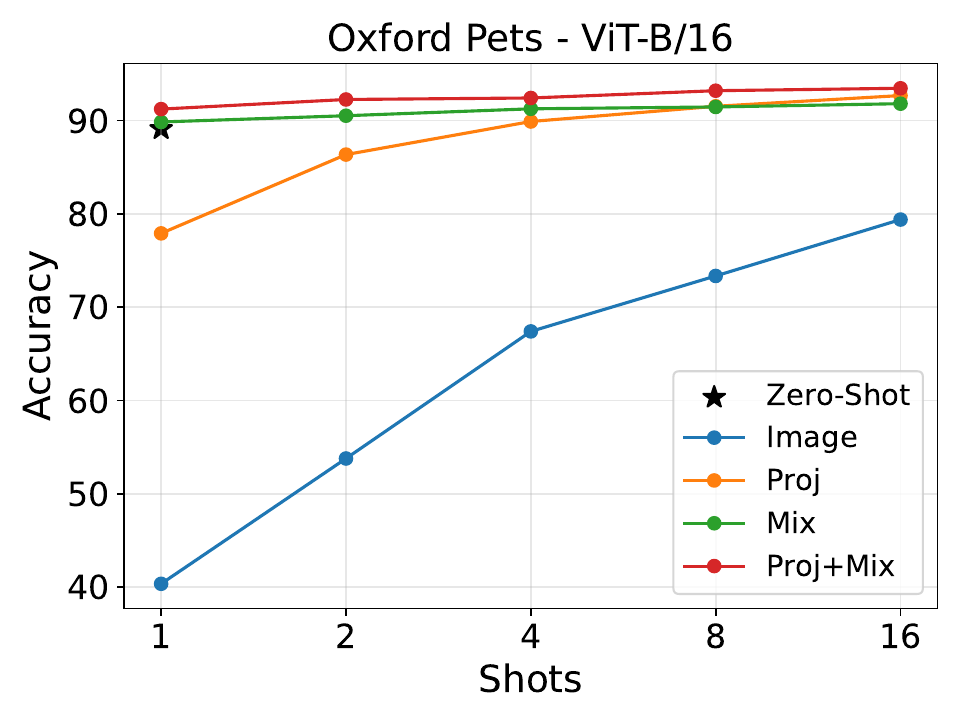}\hfill
\includegraphics[width=0.33\textwidth]{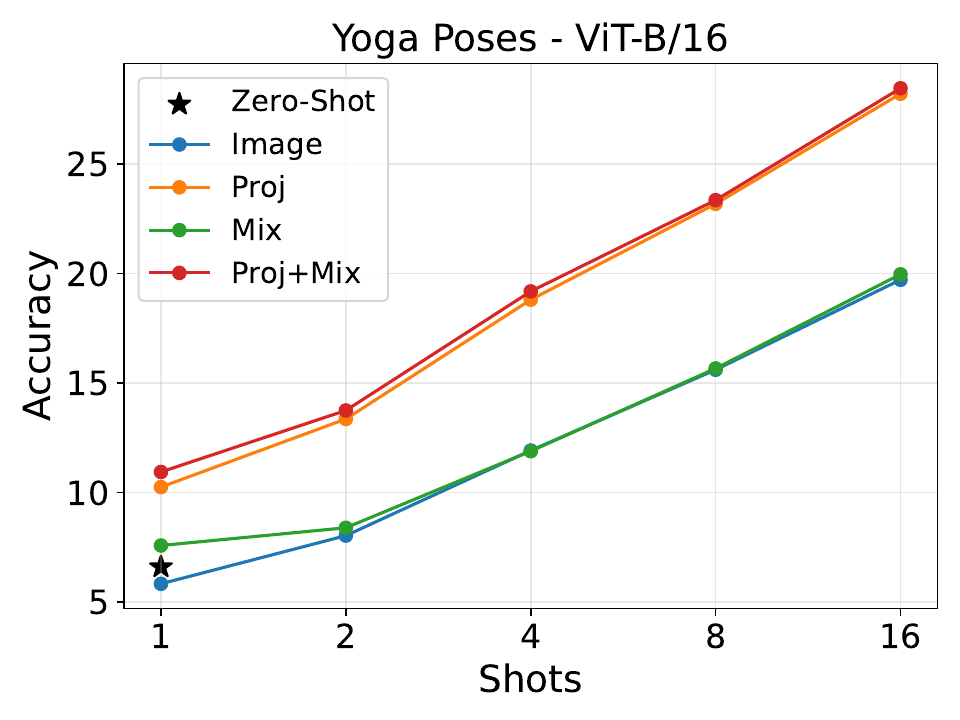}

\includegraphics[width=0.33\textwidth]{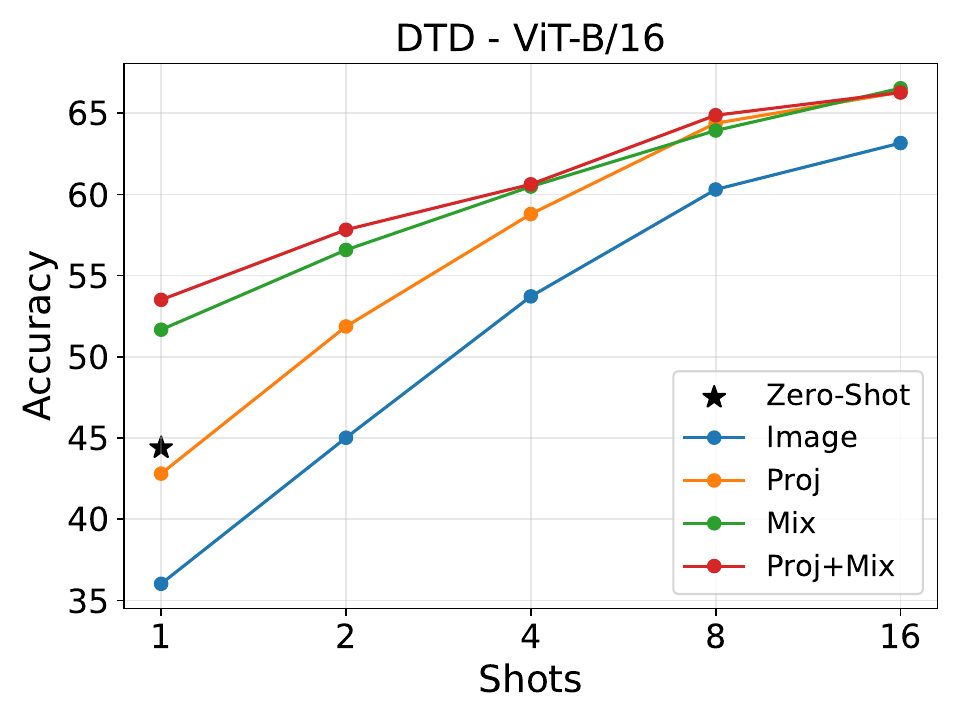}\hfill
\includegraphics[width=0.33\textwidth]{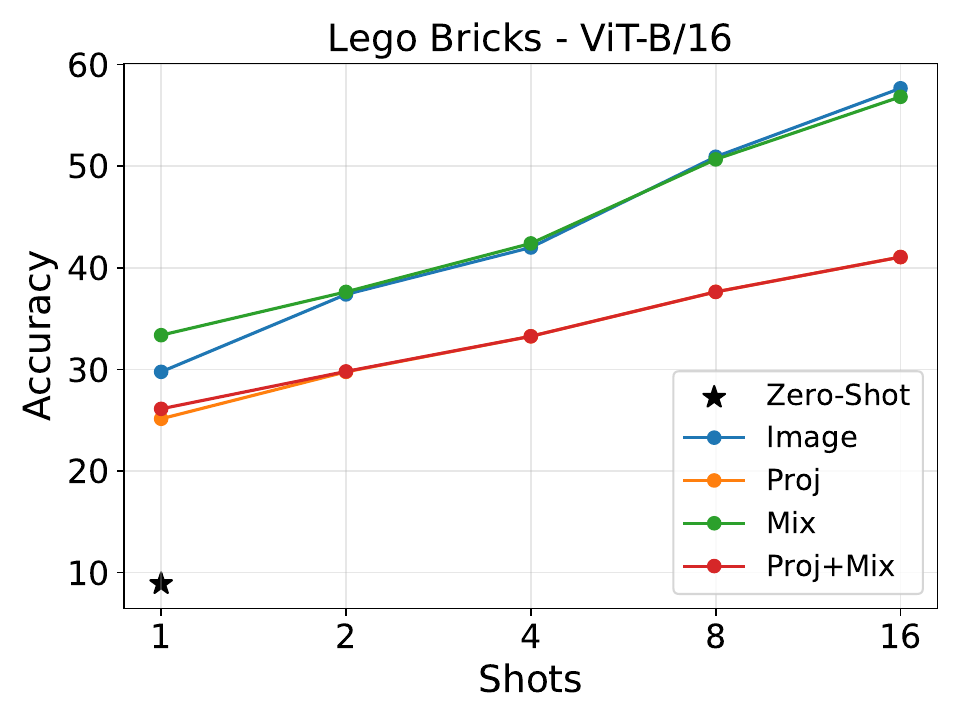}\hfill
\includegraphics[width=0.33\textwidth]{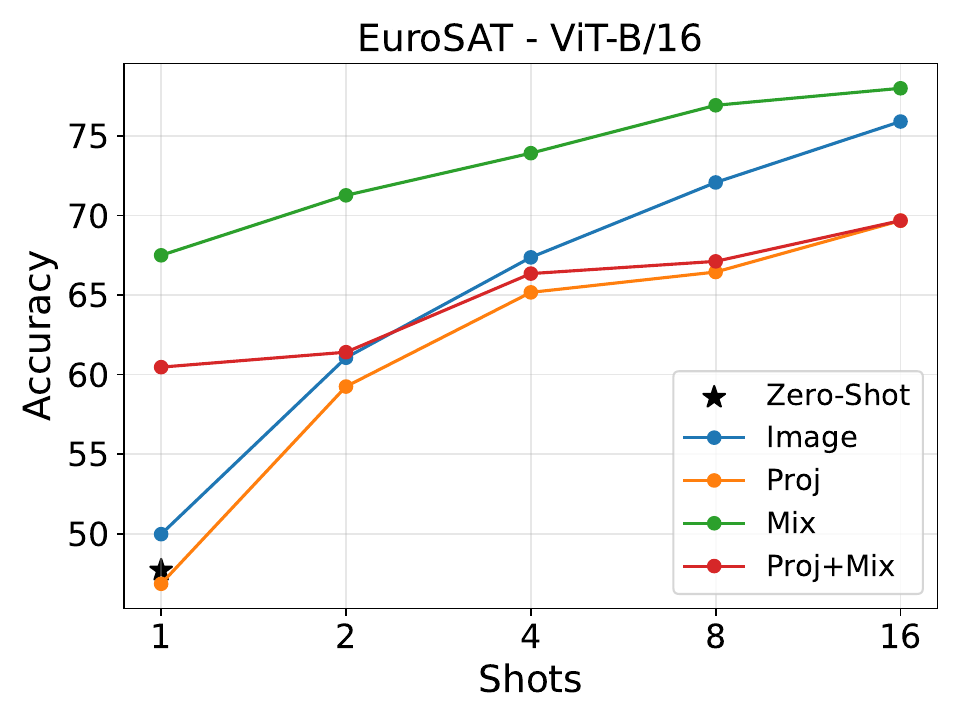}

\caption{Ablations of using NCM classifiers with image prototypes (Image), image prototypes in task-semantic subspace (Proj), mixed prototypes in the full image space (Mix) and mixed prototypes in the task-semantic subspace (Proj+Mix) on 12 datasets with CLIP-ViT-B/16.}
\label{fig:align_mix}
\vspace{-5pt}
\end{figure}

We demonstrate the impact of mixing text and image prototypes in the task-semantic subspace on 12 datasets in~\cref{fig:align_mix}. Across all datasets, we see a significant impact of NCM with naively mixed prototypes (Mix) over image prototypes (Image). NCM using the task-semantic subspace of the image prototypes (Proj) improves over NCM with image prototypes (Image) for all datasets except EuroSAT and Lego Bricks. 
Finally, we show that mixing the task-semantic component of the image prototypes with the text prototypes (Proj+Mix) improves the accuracy by non-trivial margins for most datasets across all shots.

\subsection{Performance Comparison on validation-free setting}
\label{app:valfree}

\begin{figure}[t]
\centering

\includegraphics[width=0.33\textwidth]{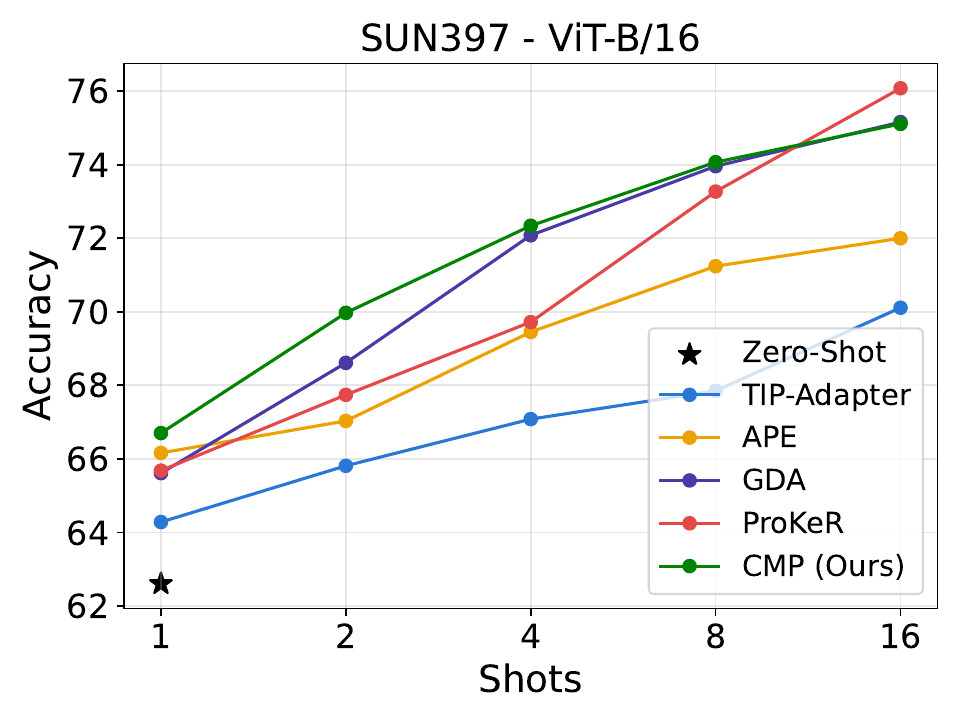}\hfill
\includegraphics[width=0.33\textwidth]{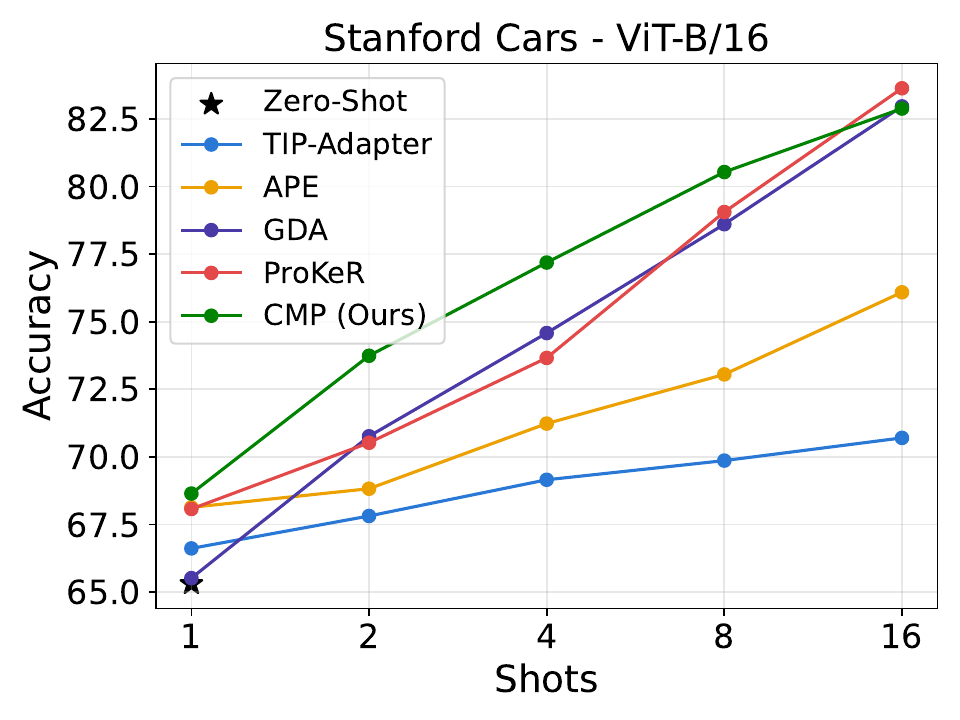}\hfill
\includegraphics[width=0.33\textwidth]{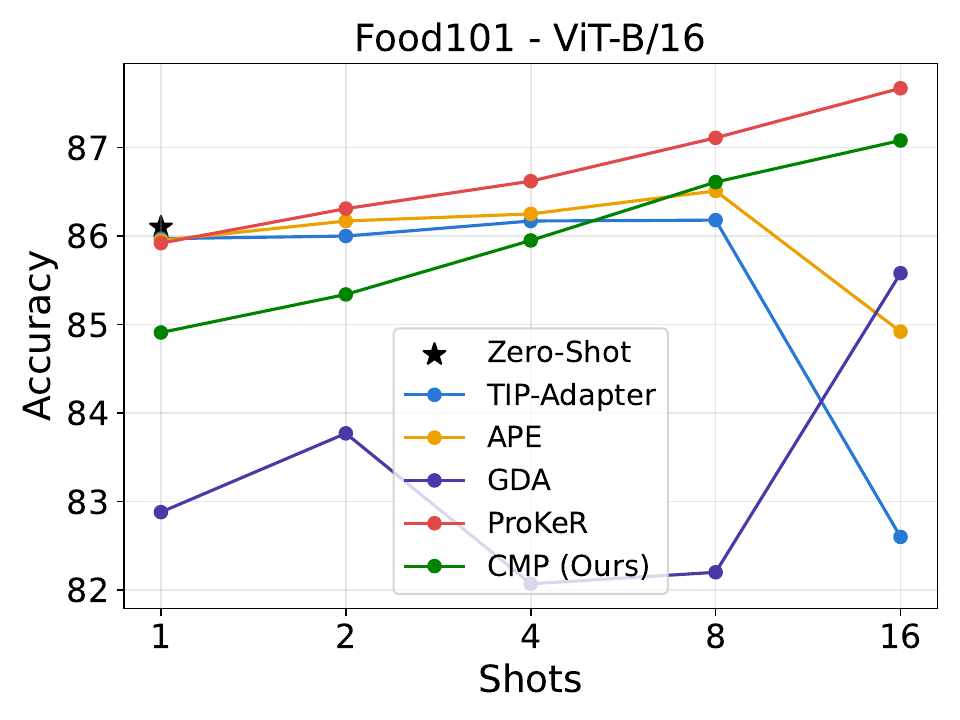}

\includegraphics[width=0.33\textwidth]{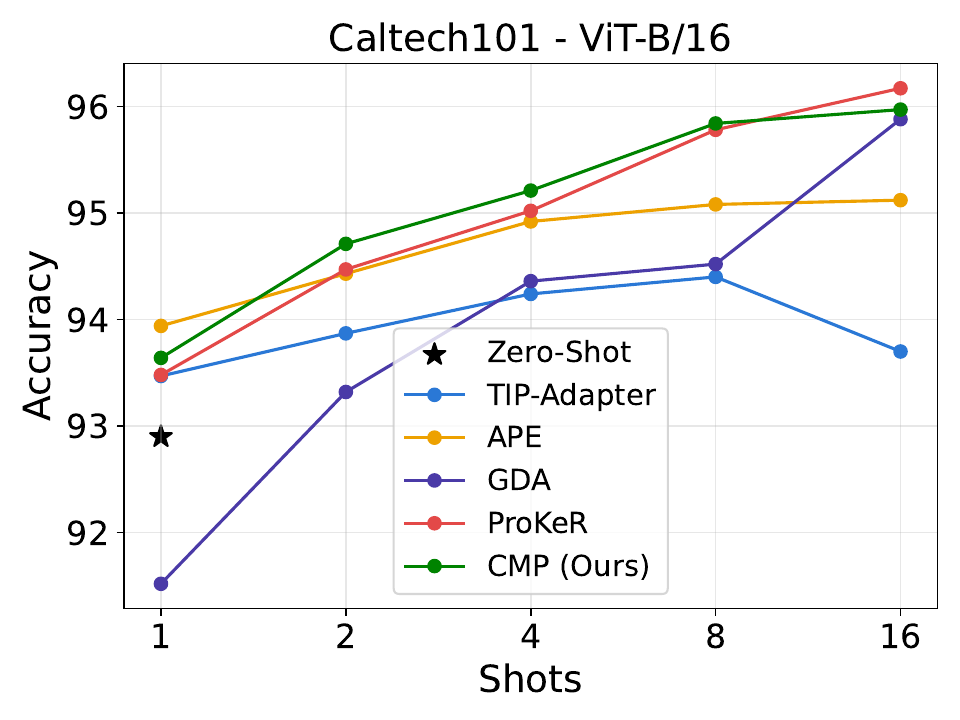}\hfill
\includegraphics[width=0.33\textwidth]{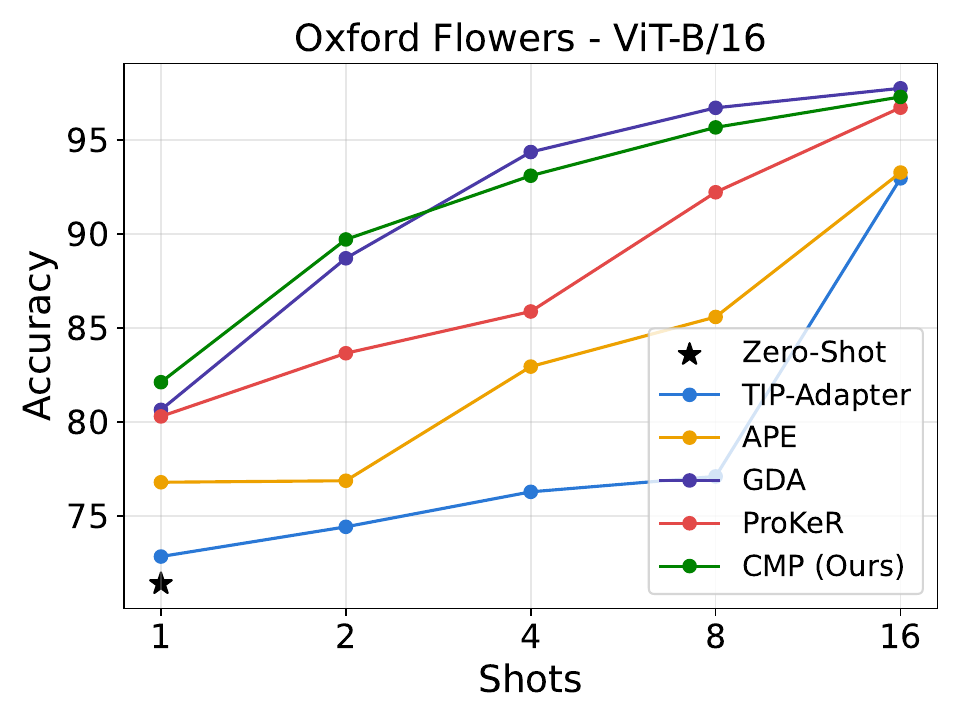}\hfill
\includegraphics[width=0.33\textwidth]{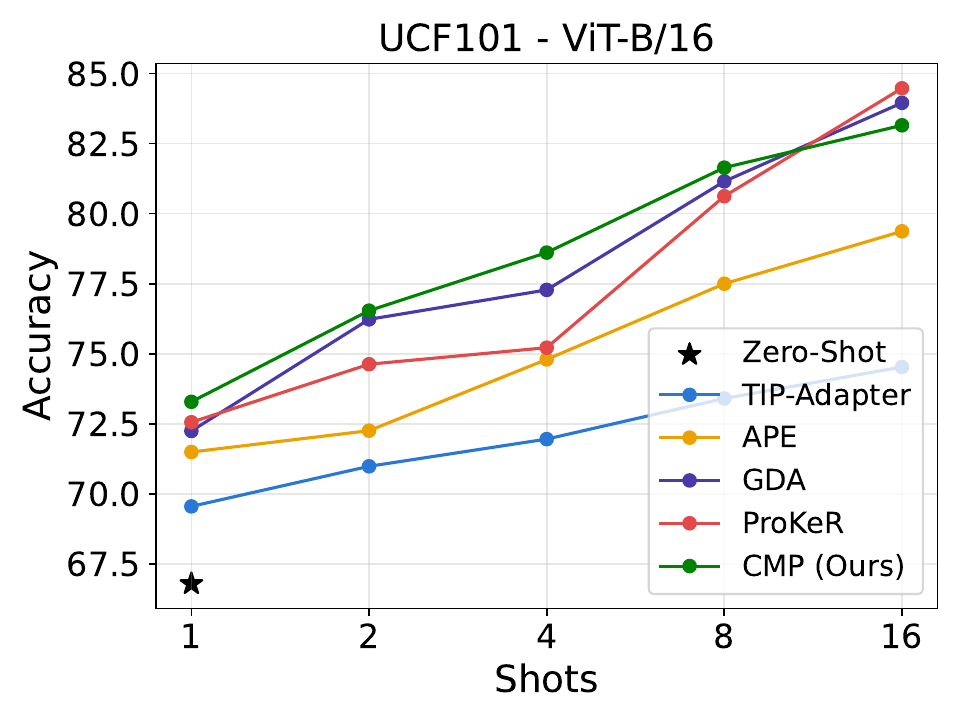}

\includegraphics[width=0.33\textwidth]{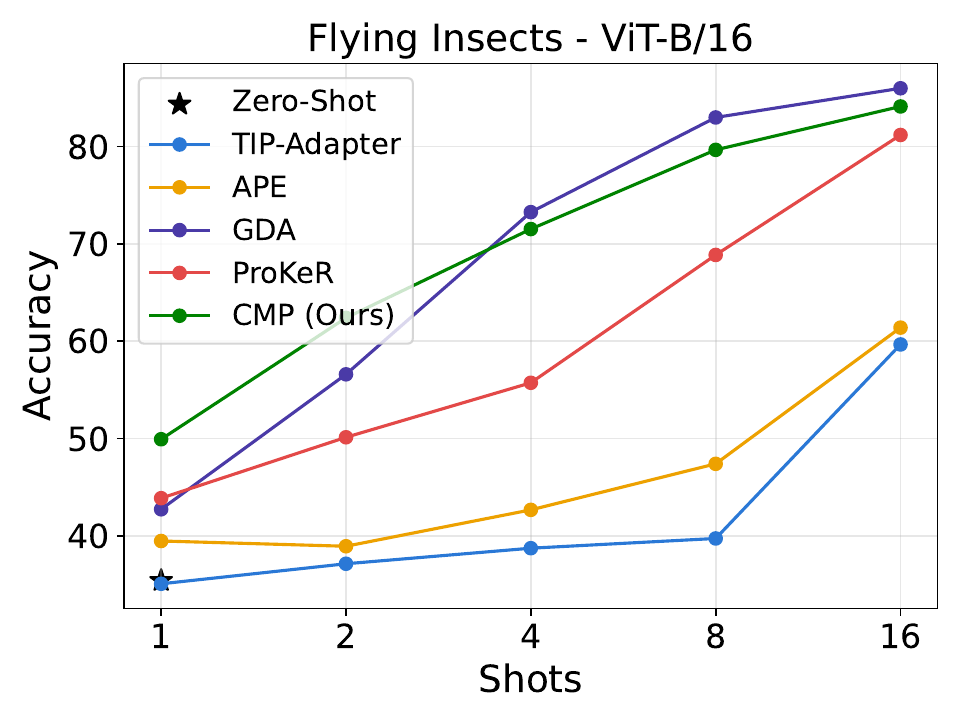}\hfill
\includegraphics[width=0.33\textwidth]{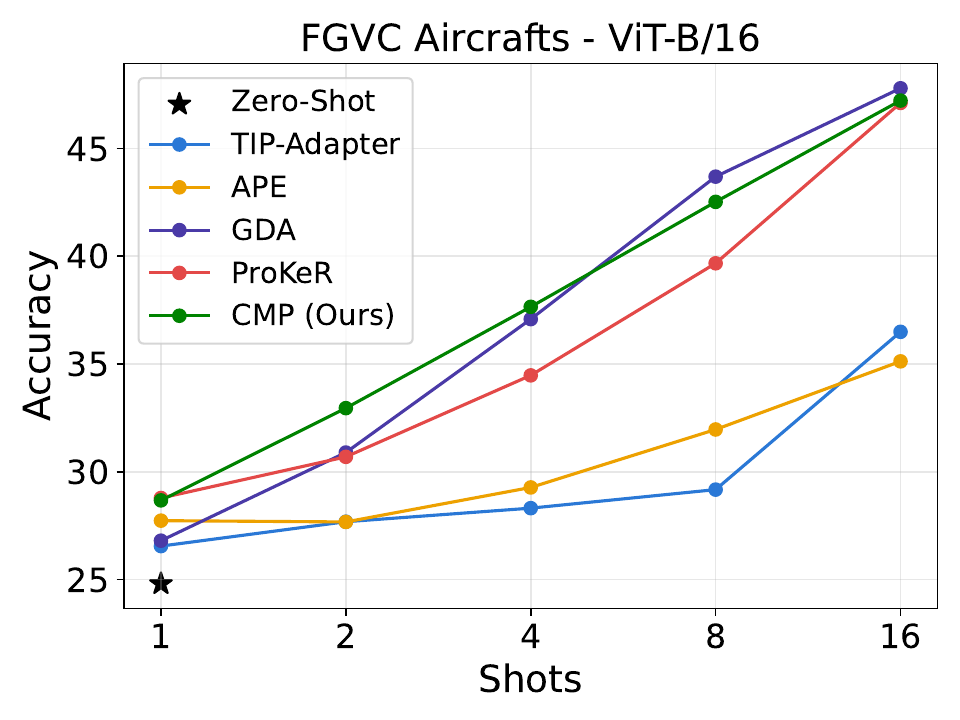}\hfill
\includegraphics[width=0.33\textwidth]{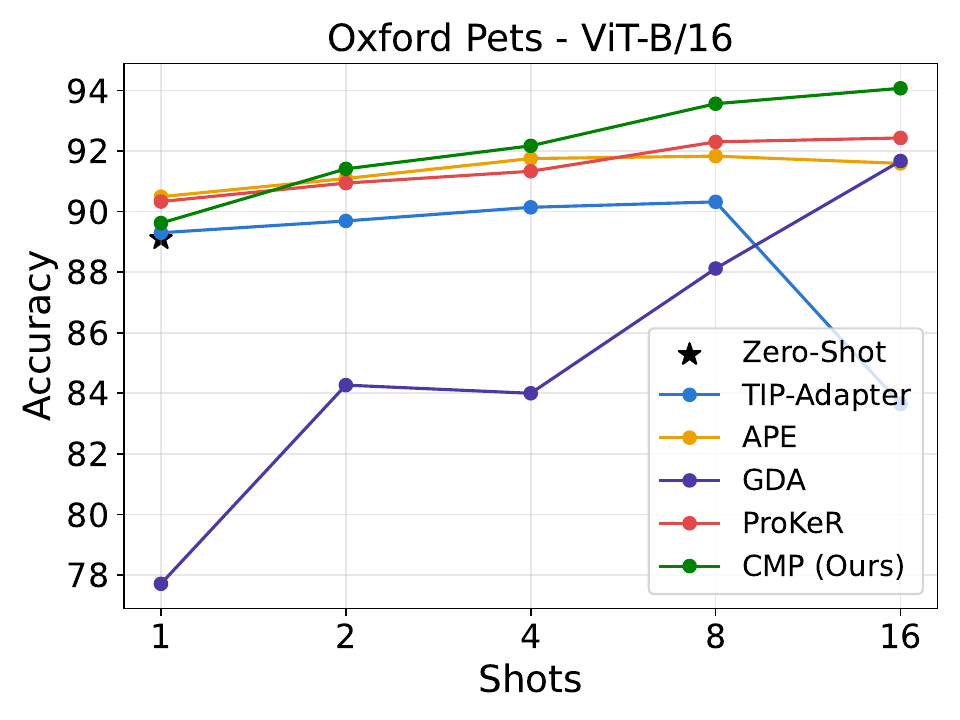}

\includegraphics[width=0.33\textwidth]{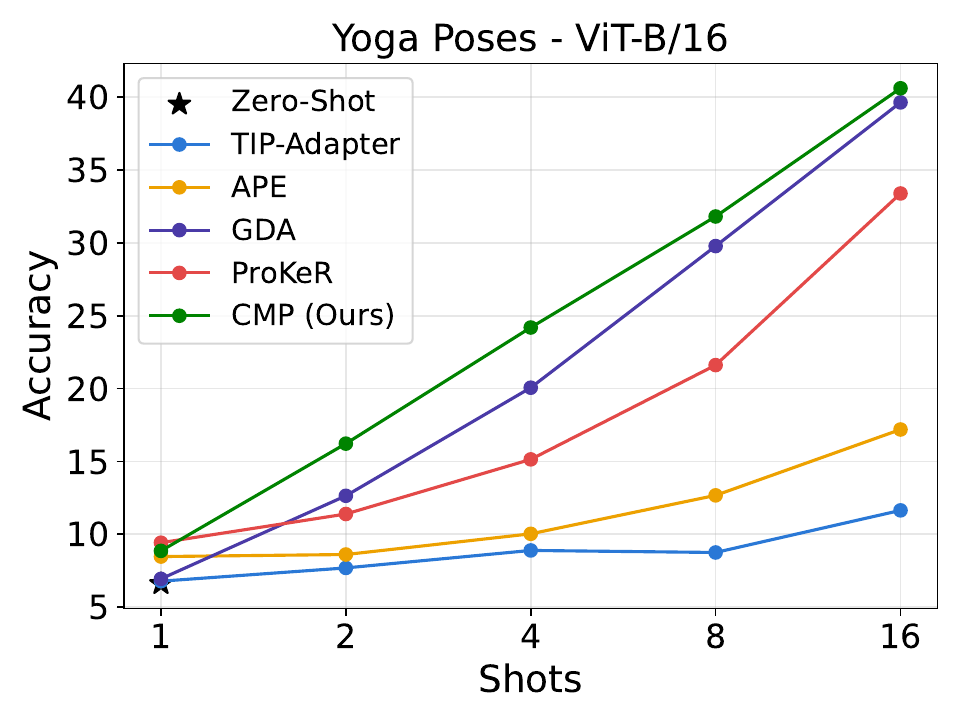}\hfill
\includegraphics[width=0.33\textwidth]{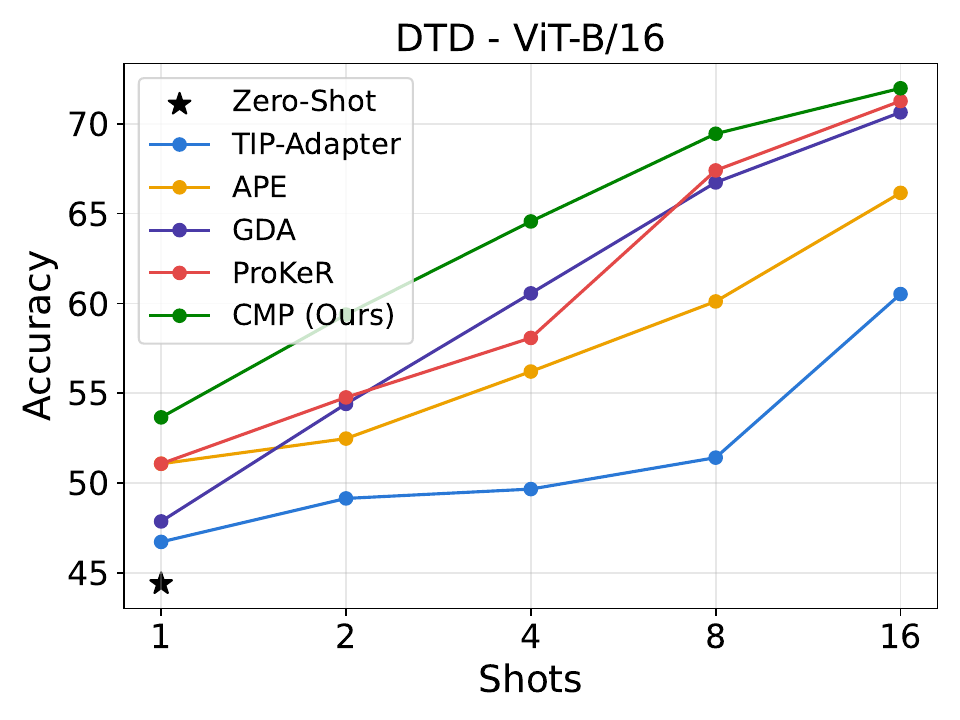}\hfill
\includegraphics[width=0.33\textwidth]{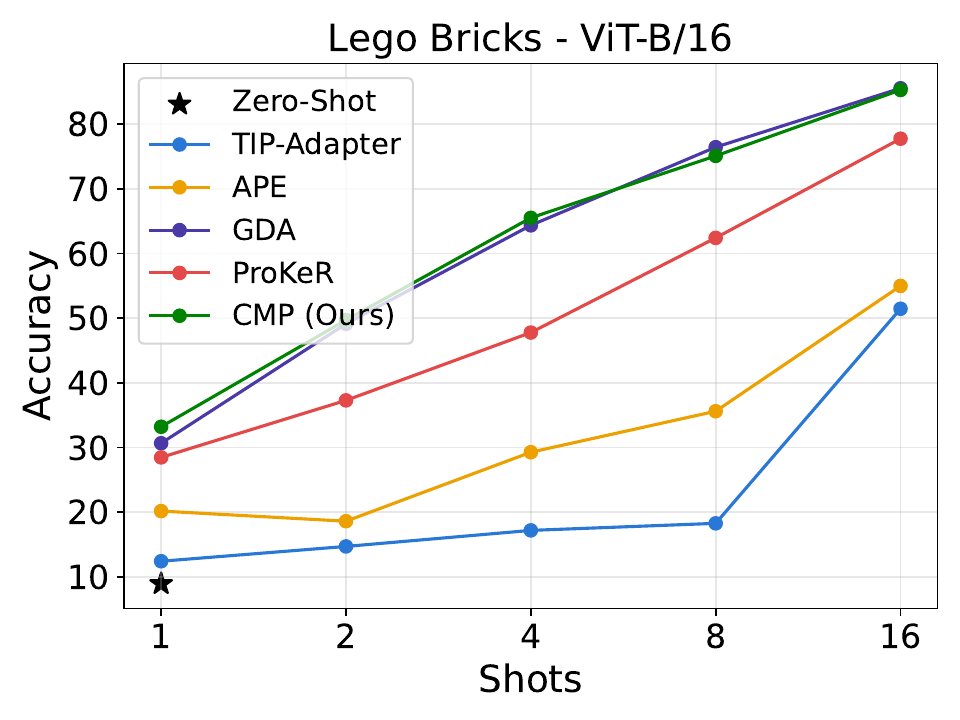}

\includegraphics[width=0.33\textwidth]{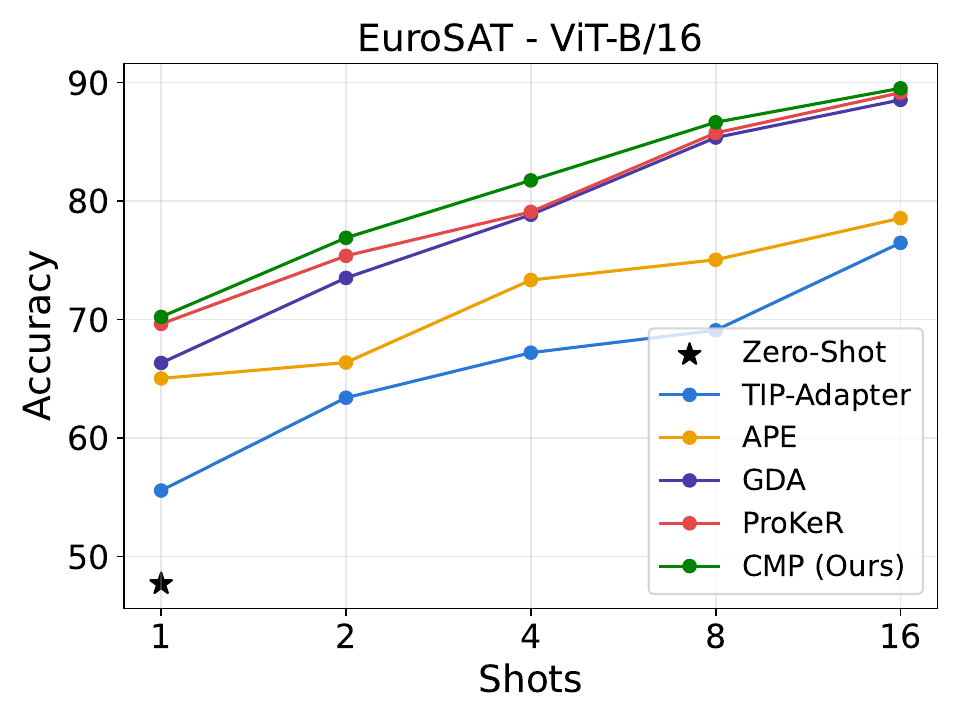}\hfill
\includegraphics[width=0.33\textwidth]{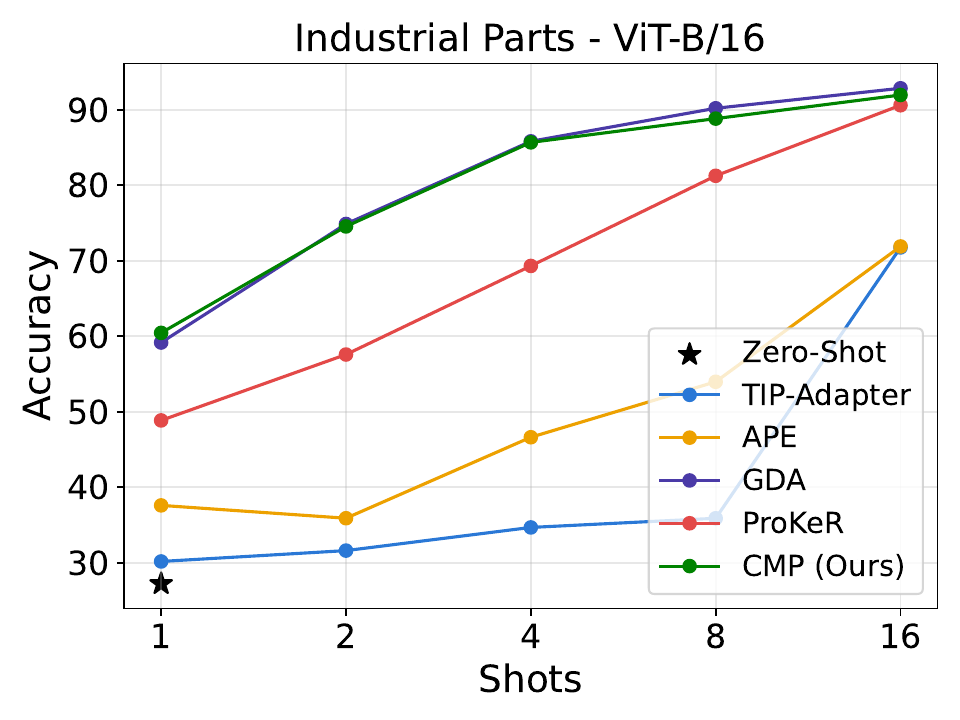}\hfill
\includegraphics[width=0.33\textwidth]{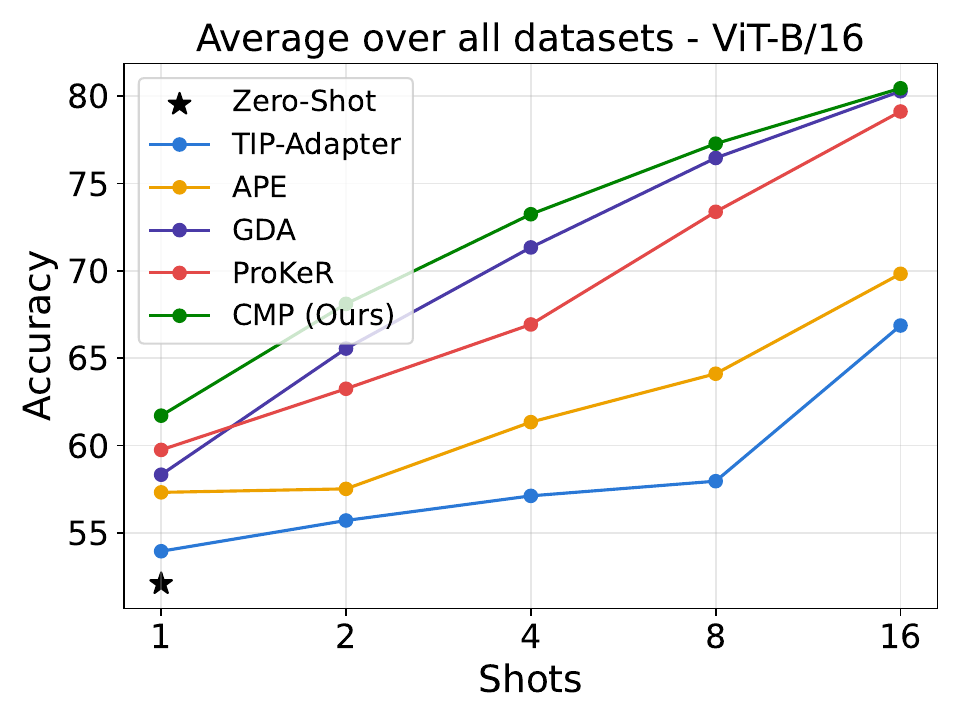}

\caption{Performance comparison of training-free methods using CLIP-ViT-B/16 in the \textit{validation-free} setting.}
\label{fig:comparison_vit}
\vspace{-5pt}
\end{figure}

We present the performance comparison with all training-free methods for 14 datasets using CLIP-ViT-B/16 in the \textit{validation-free} setting in~\cref{fig:comparison_vit}. All existing methods are very sensitive to hyperparameters and needs to carefully finetune them on a validation set. Following the validation-free setting from~\citep{bendou2025proker} where the hyperparameters are selected on the validation set of a reference dataset ImageNet and then the selected values are used for all downstream datasets. We followed this approach for all existing methods. For the proposed CMP classifier, we do not require any such hyperparameter tuning on a reference validation set. For our method CMP, we select the mixing coefficient $\lambda$ based on shots and $\gamma$ based on the cross-modal misalignment score from~\eqref{eq:ensemble} to combine the Proj+Mix (NCM) and Image (Maha) classifiers. \cref{fig:comparison_vit} shows that the proposed CMP outperforms other methods in most cases across all datasets. The performance of existing methods vary across datasets suggesting the need to carefully finetune them on dataset-specific validation sets. In contrast, CMP is relatively stable across all datasets suggesting less reliance on dataset-specific hyperparameter selection.

\begin{figure}[t]
\centering

\includegraphics[width=0.8\textwidth]{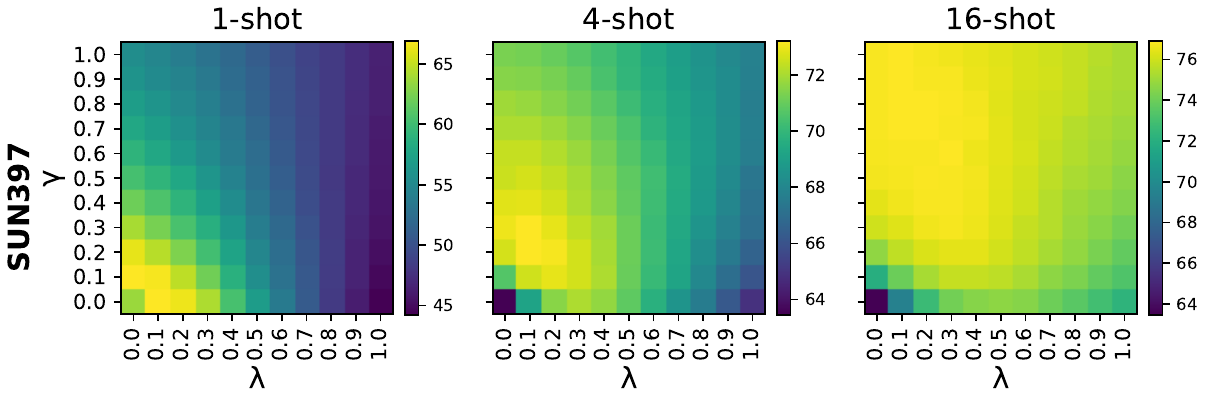}
\includegraphics[width=0.8\textwidth]{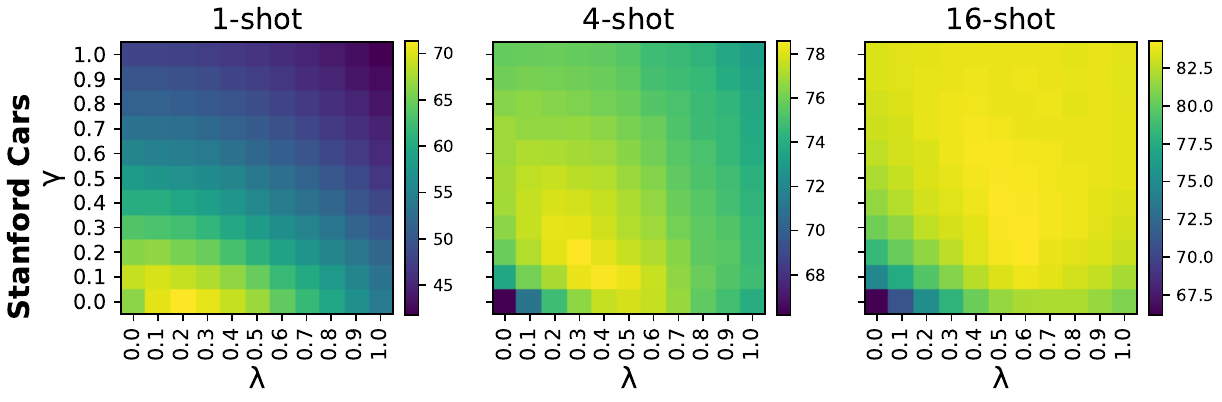}
\includegraphics[width=0.8\textwidth]{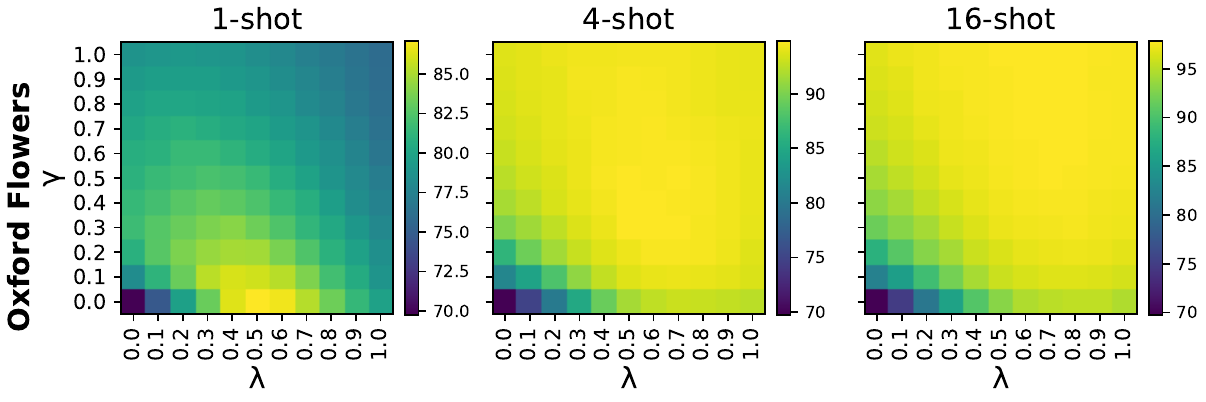} 
\includegraphics[width=0.8\textwidth]{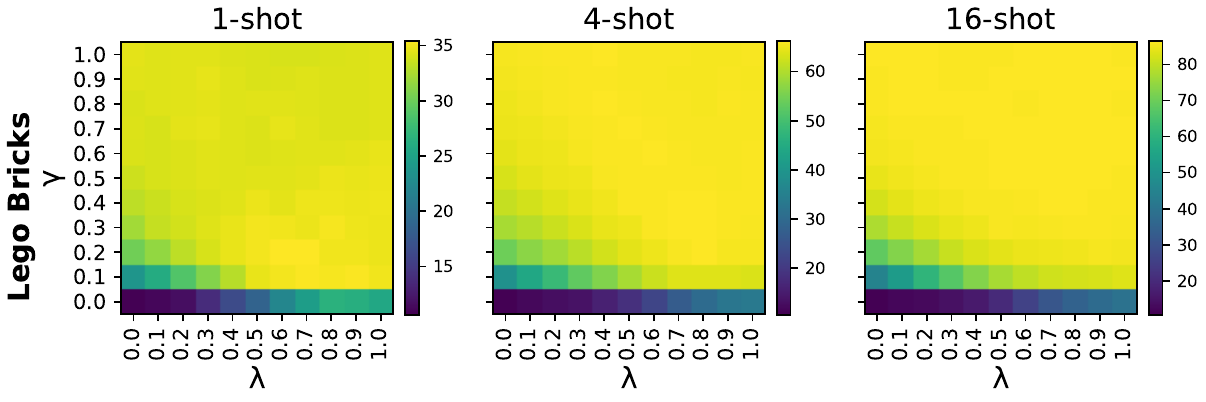}
\includegraphics[width=0.8\textwidth]{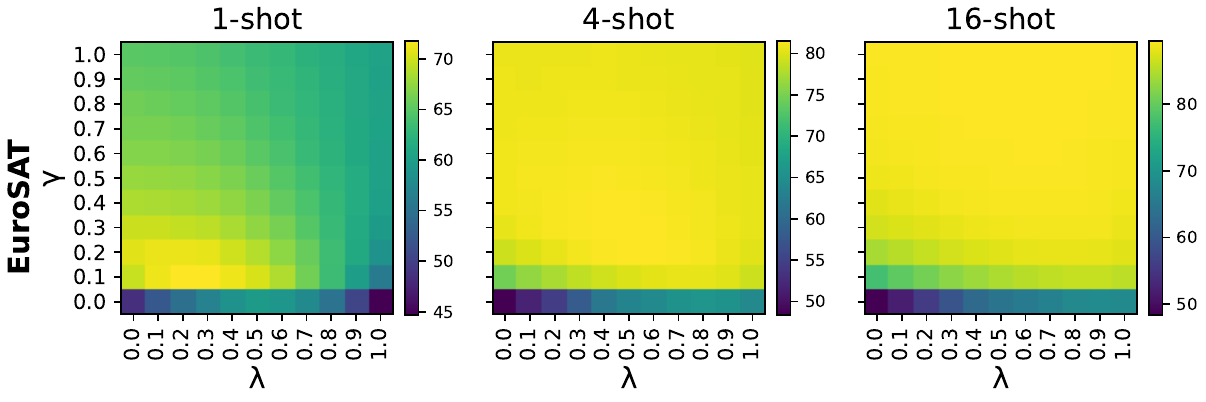}

\caption{Analysis of the sensitivity of the hyper-parameters $\lambda$ and $\gamma$ to the validation set accuracy on 5 datasets with CLIP-ViT-B/16.}
\label{fig:hyper}
\vspace{-5pt}
\end{figure}

\subsection{Analysis of Hyperparameters} 

We analyze the sensitivity of the proposed method to the hyper-parameters $\lambda$ and $\gamma$ in~\cref{fig:hyper} when they are tuned on validation sets of all datasets. 
$\lambda\in [0, 1]$ is the mixing coefficient of the Proj+Mix classifier. Higher lambda values assign higher weights to the image prototype in the task-semantic space and lower weights to the text prototype.
$\gamma\in [0, 1]$ is the weight for the image-based classifier.
For SUN397, Stanford Cars and Oxford Flowers, we see similar trends where the $\gamma$ values corresponding to higher accuracies are lower for few-shots and increases for 16-shot. This suggests that the Mahalanobis image classifier is more significant in 16-shot setting with better covariance estimates. Similarly, for $\lambda$, we see that the optimal lambda value increases with more shots where the image prototype estimates are better. Note that we always see an optimal value of $\lambda > 0$ implying the significance of mixing in the semantic subspace. For EuroSAT and Lego Bricks, we see more dependence on the image classifier with higher optimal values of $\gamma$. Similar trends across most datasets suggest that the hyperparameters are less sensitive to tasks or datasets.

\end{appendices}

\end{document}